\documentclass[letterpaper]{article} % DO NOT CHANGE THIS
\usepackage{aaai2026}  % DO NOT CHANGE THIS
\usepackage{times}  % DO NOT CHANGE THIS
\usepackage{helvet}  % DO NOT CHANGE THIS
\usepackage{courier}  % DO NOT CHANGE THIS
\usepackage[hyphens]{url}  % DO NOT CHANGE THIS
\usepackage{graphicx} % DO NOT CHANGE THIS
\usepackage{natbib}  % DO NOT CHANGE THIS AND DO NOT ADD ANY OPTIONS TO IT
\usepackage{caption} % DO NOT CHANGE THIS AND DO NOT ADD ANY OPTIONS TO IT
\usepackage{amsmath,amsfonts}
\usepackage{multirow}
\usepackage{booktabs}

\title{Rethinking Video-Language Model from the Language Input Perspective}

\author{
    Xiang Fang\textsuperscript{\rm 1}, Wanlong Fang\textsuperscript{\rm 2}, Changshuo Wang\textsuperscript{\rm 3}\thanks{Corresponding Author. }, Xiaoye Qu\textsuperscript{\rm 4},
    Daizong Liu\textsuperscript{\rm 5}
}
\affiliations{
   \textsuperscript{\rm 1}School of Software Engineering,
Huazhong University of Science and Technology\\
    \textsuperscript{\rm 2}Nanyang Technological University, Singapore\\ 
    \textsuperscript{\rm 3}University College London \\
    \textsuperscript{\rm 4}Huazhong University of Science and Technology \\
    \textsuperscript{\rm 5}Wuhan University 
     xfang9508@gmail.com,  wanlongfang@gmail.com, wangchangshuo1@gmail.com, xiaoye@hust.edu.cn, daizongliu@whu.edu.cn
}

\begin{document}

\maketitle
\begin{abstract}
Driven by the wave of large language models, Video-Language Models (VLMs) have become a significant yet challenging technology to bridge the gap between videos and texts. Although previous VLM works have made significant progress, almost all of them implicitly assume that all the texts are predefined by the specific template. In real-world applications, such a strict assumption is impossible to satisfy since 1) predefining all the texts is extremely time-consuming and labor-intensive. 2) these predefined text inputs are too restrictive and user-unfriendly, limiting their applications. It is observed that given a video input, texts with similar semantics but different templates lead to various performances. To this end, in this paper, we propose a novel plug-and-play framework  for various VLM-based methods to fully bridge videos and texts.
% text-augmented VLM method to improve video-text fusion by text rewriting. 
Specifically, we first generate positive and negative texts from the original ones  to target specific text components.  Then, we  propose an attribute-based text reasoning strategy to mine fine-grained textual semantics of generated texts. Finally, we utilize videos as guidance to conduct cross-modal bridging by designing a self-weighted loss.
% bridge videos and 
% a multi-level contrastive learning module is designed to mine the coarse-grained language information.
Extensive experiments  show that the proposed method can serve as the plug-and-play module to effectively improve the performance of state-of-the-art VLMs.
\end{abstract}

\section{Introduction}
\label{sec:intro}
% \vspace{-7pt}
Due to remarkable success, Video-Language Models (VLMs) have attracted more and more attention \citep{rizve2024vidla,liu2023exploring,wang2025taylor,fang2026towardsicml,kuai2026dynamic,wang2025point,fang2025your,zhang2025monoattack,fang2023hierarchical,liu2024towards,yang2025eood,fang2022multi,fang2026cogniVerse,lei2025exploring,fang2023you,wang2025dypolyseg,fang2025hierarchical,yan2026fit,fang2025adaptive,wang2026topadapter,cai2025imperceptible,fang2026slap,wang2026reasoning,fang2026immuno,wang2026biologically,fang2026disentangling,wang2025reducing,fang2026advancing,fang2026unveiling,wang2026from,liu2023conditional,liu2026attacking,wang2025seeing,fang2026towards,fang2025multi,fang2024fewer,liu2024pandora,fang2024multi,fang2025turing,fang2024not,liu2023hypotheses,fang2024rethinking,liu2024unsupervised,fang2023annotations,xiong2024rethinking,fang2021unbalanced,wang2025prototype,zhang2025manipulating,fang2026align,tang2024reparameterization,fang2025adaptivetai,tang2025simplification,fang2021animc,cai2026towards,fang2020v}.  VLMs require cooperation from both computer vision and natural language processing for precise semantic alignment and have a wide range of applications such as video summarization \cite{abdar2024review,fang2020double} and video question answering \cite{yu2024self}. Benefiting from the strong knowledge integration ability in large language models (LLMs) \cite{fang2026align,zhang2025can,liu2023exploring,liu2024unsupervised}, VLMs show superior performances in solving complex image-language tasks by utilizing appropriate human-instructed prompts \citep{hakim2023leveraging,wang2025point,wang2025reasoning,wang2025dypolyseg,wang2026biologically,wang2025taylor}. 
% Since the real-world videos contain much temporal information, LVLMs still have difficulty to handle the real-world videos. Besides, 
In VLMs, the sentence text is the most important input that accompanies the video due to its human-friendly and descriptive nature \cite{zhang2018better,zhang2022costa}.

 \begin{figure*}[t!]
  \centering
    \includegraphics[width=\textwidth]{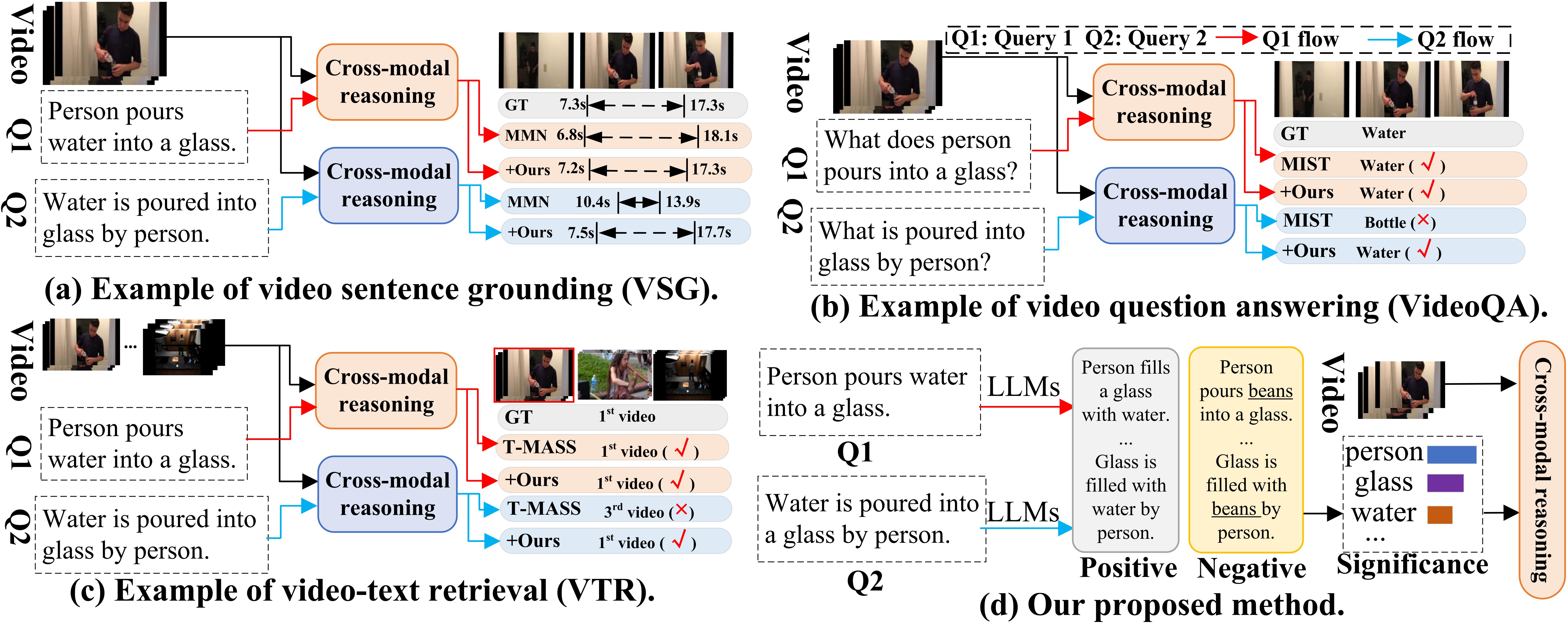}
    % \vspace{-10pt}
    \caption{(a-c) Example of the VLM tasks (VSG, VideoQA and VTR), where our proposed method can serve as a plug-and-play module for previous VLM models to enhance their efficiency. (d) Pipeline of our  method.}
    % \caption{Performance comparison between existing approaches and our proposed method in video-language joint tasks. Here, conventional models fail to properly attend all parts of the sentence, \textit{e.g.}, \textbf{\textit{shot}}. This is alleviated by the incorporation of targeted hard samples using our proposed mechanism.}
    \label{fig:intro}
% \vspace{-15pt}
\end{figure*}

% As a significant and challenging multimedia task, 
% Video Sentence Grounding (VSG) aims at retrieving the start and end timestamps of the target video moment in an untrimmed video semantically according to a sentence text. As shown in Figure~\ref{fig:intro}(a), VSG requires cooperation from both computer vision and natural language processing for precise semantic alignment and has a wide range of applications such as video summarization and video question answering. In the VSG task, the sentence text is the most important input that accompanies the video due to its human-friendly and descriptive nature.

Current VLMs contain three main popular yet challenging tasks in  Figure \ref{fig:intro}: video question answering (VideoQA) \citep{gao2023mist}, video sentence grounding (VSG) \citep{zhang2023temporal} and video-text retrieval (VTR) \citep{zhu2023deep}. VideoQA is a significant multi-modal task where a model is given a video along with a natural language question about the video content, and it must generate or select the correct answer. The task requires the model to understand the visual cues in the video, as well as the language of the question, to provide relevant and accurate responses. Given a language text and an untrimmed video, VSG aims at retrieving the start and end timestamps of the target video moment, semantically according to a sentence text.
Given a language text, VTR targets to retrieve relevant videos from a large video database, which can be either text-based (text-to-video retrieval) or video-based (video-to-video retrieval). The significant goal of VTR is to find videos that best match the given input by analyzing visual content, actions, and sometimes audio cues.
The  performance in above three downstream tasks depends on their capability to extract video features and align them with text features. Some VLM-based methods only perform data augmentations on the video input to improve the model robustness during training in every epoch. Unfortunately, existing methods only utilize predefined texts without any augmentation. In real-world applications, these sentence texts with similar textual semantics might be inputted with different structure/vocabulary variations from various users. As shown in  Figure \ref{fig:intro}(b), the text (``Person pours water into a glass'') shares the same semantics as the text (``Water is poured into a glass by person'').
% However, when we directly utilize them as the textual inputs in state-of-the-art methods, these methods obtain significantly different grounding results, which is unreasonable. 
% their performance will drop seriously, as shown in Figure \ref{xxx}. 
% Although two texts have similar semantics, 
However, previous methods yield dissimilar grounding results in  Figure \ref{fig:intro}. 
The main reason is that these methods cannot utilize their weak text encoder to learn discriminative textual representations, which illustrates the significance of handling the text variations.
Therefore, it is important to ensure that the designed VLM-based method is robust enough to deal with various texts with different templates. However, existing language augmentation approaches are not sufficiently effective to integrate the multi-modal inputs. Some methods target to replace or mask some words in a sentence, which only brings limited influence in diversifying the text structure/vocabulary. It is comparatively weaker than video augmentations. The target language augmentation approach should effectively rewrite sentence texts while reserving the core textual semantics. The approach is urgently required for model training to achieve the best  results.

In this paper, we propose a simple yet highly effective framework to improve the robustness and performance of VLMs.
Specifically, we  generate multiple variants of each text in the video-text pairs. Based on the variation-origin pairs, we utilize them as examples to diversify all the texts in video-text datasets. Different from previous sentence augmentation works that only change some words to preserve sentence structures, we generate rich variations for diverse text inputs due to their extensive training datasets and emergent properties.  Based on the above sentence augmentation, each video corresponds to diverse texts. Moreover, we utilize the original as the anchor to generate various hard negative texts  by changing different sentence parts. In particular, we utilize precise prompt engineering to modify specific parts of the sentence with the rest parts unchanged. Also, we generate positive samples that lie relatively far from the anchor in the embedding space. To further understand the latent textual semantics, we design an attribute-based text reasoning strategy for fine-grained text mining.
% To analyze the relative significance of each sentence part, 
To bridge videos and texts,
we incorporate these generated text samples with the video as  guidance by a self-weighted cross-fusion loss.
% to conduct self-weighted cross-modal bridging.
% by a weighted contrastive loss function.
With these diverse texts, we target to train VLM models with augmentation from the text perspective. 
% For convenience, we randomly choose a text augmentation from multiple diverse texts.

% generate two categories

%=====================================================

Our main contributions are summarized as follows: 1) We make the first attempt to explore the effect of the template-free text for the robust VLM task, where we localize the target activity by a user-friendly text with any form instead of a predefined text.   Also, we propose a novel plug-and-play framework for VLM-based methods to fully understand the text inputs.
 % generate positive and negative texts, where each negative text is used to highlight a sentence component. 
2) We first generate diverse positive and negative texts by augment the original text from  word  and structure levels.
% To obtain diverse positive and negative texts, we augment the text by both word-level and structure-level for rewriting texts.
To selectively integrate these generated texts, we design two modules (generating multi-level texts module and attribute-based text reasoning module) to understand the text input from different granularity.  Besides, a video-guided self-weighted cross-modal bridging loss is proposed to integrate these sentence components by assigning adaptive weights to these components. 3) For three representative downstream tasks (VSG, VideoQA and VTR), we conduct experiments on many popular yet challenging datasets. To obtain different text types, we leverage LLMs to construct a small set of variation-origin by two strategies:  original datasets and augmented datasets. 4) Extensive experimental results on both original datasets and augmented datasets show that 
    % our proposed model outperforms existing approaches by a large margin. Moreover, 
    our proposed model can serve as a plug-and-play module for state-of-the-art VLM-based methods.

\section{Related Works}\label{sec:related_works}

%===========================================
\begin{figure*}[t]
  \centering 
% \vspace{-25pt}
 \includegraphics[width=\textwidth]{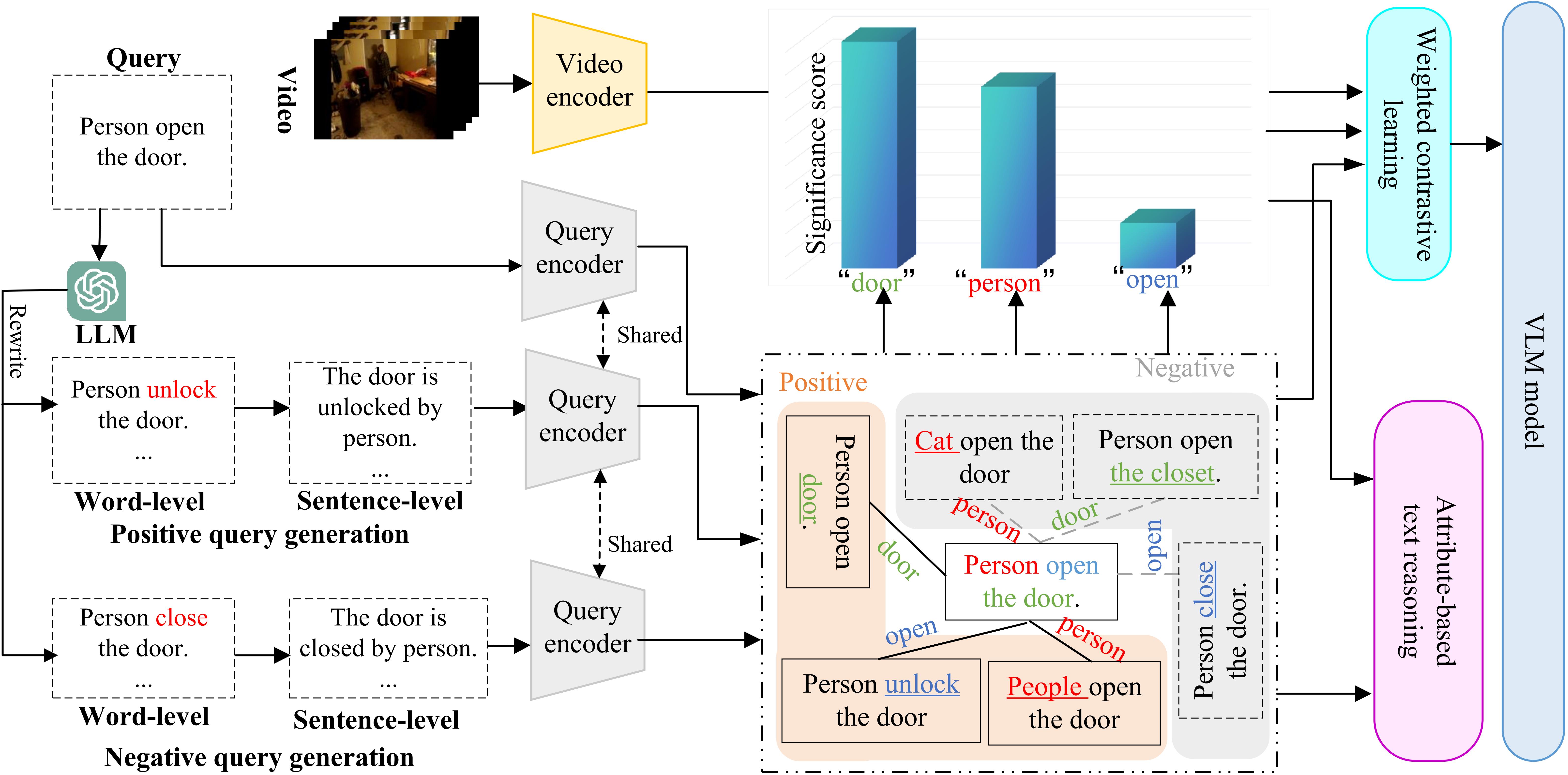}
 % \vspace{-8pt}
   \caption{ Illustration of our proposed framework. }
   \label{fig:pipeline}
  % \vspace{-15pt}
\end{figure*}
%===========================================
% \textbf{Large vision-language models.} The breakthrough of LLMs in language-oriented tasks \citep{ma2024survey,du2024multi} and the emergence of GPT-4 have prompted researchers to explore the potential of LLMs in assisting with a range of tasks across multi-modal scenarios \citep{carolan2024review,yin2024lamm}. This has led to the development of a new field, namely large vision-language models (LVLMs). A variety of strategies and models have been proposed to address the discrepancy between text and other modalities. Some works employ learnable texts to extract visual information and generate language using LLMs conditioned on the visual features. Models including GPT-4o, MiniGPT-4 and LLaVA  learn simple projection layers to align the visual features from visual encoders with text embeddings for LLMs. Additionally, parameter-efficient fine-tuning is adopted by introducing lightweight trainable adapters into models. Several benchmarks have verified that LVLMs demonstrate satisfactory performance on visual perception and comprehension.

\textbf{Video-language models (VLMs).} The breakthrough of LLMs in language-oriented tasks \citep{ma2024survey,zhang2025can} and the emergence of GPT-4 have prompted researchers to explore the potential of LLMs in assisting with a range of tasks across multi-modal scenarios \citep{carolan2024review}. This has led to the development of a new field, namely VLMs. A variety of strategies and models have been proposed to address the discrepancy between text and other modalities. Some works employ learnable texts to extract visual information and generate language using LLMs conditioned on the visual features. Models including GPT-4o, MiniGPT-4 and LLaVA  learn simple projection layers to align the visual features from visual encoders with text embeddings for LLMs. Additionally, parameter-efficient fine-tuning is adopted by introducing lightweight trainable adapters into models. Several benchmarks have verified that VLMs demonstrate satisfactory performance on visual perception and comprehension.

Although these methods have achieved promising results, all of them heavily rely on correctly aligned multi-modal datasets.
Therefore, it is highly expected to develop a VLM model that is robust to different texts with similar semantics, which has not been studied as far as we know. Thus, \textit{we make the first attempt to reveal the text understanding problem in VLM task and propose to eliminate the negative impact of the different texts with any template.} 
% \textbf{The full related work section is placed in supplementary material.}

\section{Methodology}
\label{sec:methodology}
We elaborate on the proposed method, 
% \textit{i.e.}, xxx, 
which strengthens the text encoder to obtain consistent representations for various semantically similar texts in real-world  multi-modal datasets (\textit{e.g.}, Charades-STA \citep{sigurdsson2016hollywood}), multiple semantics-similar texts often share a video moment with the target activity. For example,  ``Person opens the door'' and ``The door is opened by person'' have similar semantics. Since the text template is fixed, it is still challenging to diversify the text input. Thus, we design a text augmentation module to generate semantically similar texts.
% . We can treat them as semantically similar texts when we train the designed VLM-based model. Since the dataset scale is limited and the text template is fixed, it is still challenging to diversify the text input. Thus, we design a text augmentation module to generate semantically similar texts.
% learns the uncertain and fine-grained cross-modal alignment in WS-VTG task. 
% Specifically, our method conceptualizes the interaction between the video and text as a game, where each frame/word acts as a player and forms coalitions. 
% During the game learning, visual frame and linguistic word representations with strong semantic relevance will cooperate to form a new coalition, the benefits of which are quantified through the game interaction index. Our method maximizes payoff by strategically forming coalitions to achieve fine-grained semantic alignment between video and text.
The overall framework is shown in Figure \ref{fig:pipeline}.
% , where we first conduct self-modal game to enhance the self-modal instance semantics and then devise multi-level cross-modal games to learn the fine-grained and uncertain cross-modal alignment. At last, we predict the segment boundary based on the aligned frame-wise scores.

\noindent \textbf{Problem statement.}
% Given multi-modal training set $\{\mathbb{V}_n, \mathbb{Q}_n, \mathbb{Y}_n\}_{n=1}^N$, each untrimmed video $\mathbb{V}_n$ is represented as $\mathbb{V}_n=\{v_{n,t}\}^{T}_{t=1}$ frame-by-frame, where $v_{n,t}$ is the $t$-th frame of $n$-th video and $T$ is the number of total frames. Similarly, the sentence text $\mathbb{Q}_n$ with $M$ words is denoted as $\mathbb{Q}_n=\{q_{n,m}\}^{M}_{m=1}$ word-by-word. 
% Given a video-text pair $({V}=\{v_{t}\}^{T}_{t=1},{Q}=\{q_{j}\}^{J}_{j=1})$, VSG aims to retrieve a specific moment ${Y}=(t^s, t^e)$ starting at the timestamp $t^s$ and ending at timestamp $t^e$ in the video $V$, which corresponds to the same semantics as text $Q$, where $v_{t}$ is the $t$-th frame in video $V$ and $q_{j}$ denotes the $j$-th word in text $Q$. 
Due to the strong language processing ability of of LLMs, we utilize LLMs to generate various texts by replacing different components for simulating the practical labeling process in the format-free setting.
% To simulate the practical labeling process in the format-free setting, considering that the pre-trained LLM can provide much textual knowledge for text generation, we utilize LLM to generate various texts by replacing different components.
% We denote $\{Q_1, \dots, Q_M\}$ as the textual input set in the VLM task, where $M$ denotes the total number of sentences.
% in the training set, which is the textual input of our model. 
Previous VLM methods \citep{yu2023self,wang2022negative,wang2024text} cannot well handle these texts with similar semantics since they do not fully understand the textual information in the sentence. 
% To better model the semantic relationship between different text components with videos, we target these text components that are later used for training. Moreover, we leverage a pre-trained LLM to generate these negative and positive samples with the help of extensive prior language knowledge in LLM.
% \noindent \textbf{Pipeline.}
To address the existing models' limitations in correlating major sentence parts with suitable video representations, we present a novel plug-and-play framework to fully understand the language input by generating negative and positive samples targeting specific sentence parts. 
These samples facilitate improved perception of specific parts of the sentence, eventually enhancing the understanding of video-language correlation. We use the generated samples as auxiliary samples alongside the original training samples by employing a novel video-guided self-weighted cross-modal bridging loss. The proposed approach is application-agnostic and can be adopted successfully in various downstream multi-modal tasks. 
% Our approach is summarized in Figure \ref{fig:general_Figure}. 

% This section is organized as follows: We outline our sample generation procedure in Section \ref{subsec:sample_gen}. Followed by this, Section \ref{subsec:cons_loss} describes the procedure of applying our proposed contrastive loss by incorporating generated samples. In Section \ref{subsec:weight_module}, we introduce our importance estimation module that adaptively weighs different sentence components based on their saliency. Finally, Section \ref{subsec:va_attn} provides the details of different video-language tasks on which we evaluate our method.

\subsection{ Multi-level Text Augmentation}
\label{subsec:positive}
% \subsection{Generating Multi-level Texts for Coarse-grained Language Alignment}
% \label{subsec:positive}
By treating original texts as anchors, we  leverage LLMs for
generating positive and negative texts to fully understand the texts, where we regard the generated text sharing similar semantics as a positive text; otherwise, the generated text is negative.
% To obtain these semantics-similar texts, for each original text, we generate text augmentation for the positive texts. 
Real-world  datasets include multiple video-text pairs $(V, Q)$, where $V$ denotes the video and $Q$ denotes one of corresponding texts.
We denote $\{Q_1, \dots, Q_M\}$ as the textual input set in the VLM task, where $M$ denotes the total number of sentences.
For the $m$-th text, we denote the generated positive text as $P_m$ and the negative text as $N_m$ .
% , where we use the positive text $p_i$ rather than the original text as the textual input. 
To obtain diverse generated texts, we adopt three text augmentation approaches: human-based , chat robot-based  and  open-source LLM-based.
% Therefore, we will generate various text augmentations by LLMs in Section \ref{subsec:sample_gen}. 
% Inspired by the effectiveness of In-context Learning \citep{xxx}, we firstly utilize LLM to condition on some instance, and then make predictions. To utilize the ICL strategy for positive text generation, we generate multiple text augmentations, which will be employed as instances in the prompt context. 
For convenience, we term the pair of generated text and original text as variation-origin text pairs. 
% Thus, we generate these variation-origin  pairs by the following approaches:
 These texts differ in two levels: word level and sentence structure level.
Therefore, we have two types of text augmentation by a two-step  process: word-level augmentation and structure-level augmentation.
For convenience, we take the positive text augmentation as an example.

\noindent \textbf{Word-level  augmentation.} In the first step, we directly rewrite the original text by changing some words.
% based on the augmentation methods in Section \ref{subsec:sample_gen}. 
% The real-world multimedia applications always contain a lot of noise (\textit{e.g.}, superfluous punctuation). 
% Since original texts always contain much noise (\textit{e.g.}, unnecessary superfluous punctuation), we treat the word-level rewriting step as responsible for converting these noisy texts into a more straightforward text template, which is common in common language usage. 
With pre-trained LLMs, we can rewrite all the texts by the following prompt:
$P'_{i,m} \leftarrow \texttt{LLM} (Q_m,$ ``Rewrite the text `\textbf{text}' concisely by \underline{changing the $i$-th word} while keeping the meaning''),
% \small
% \begin{equation}
% \begin{split}
%     P'_{i,m} \leftarrow \texttt{LLM} (Q_m,\text{``Rewrite the text `\textbf{text}' concisely by \underline{changing some words} while keeping the meaning''}),
%      \end{split}
% \end{equation}\normalsize
% \textit{``Rewrite the text `\textbf{text}' concisely \underline{by changing some words} while preserving the meaning.''}, 
where \textbf{text} is substituted with the given text, and we utilize the underlined text to prompt our model for producing morphologically diverse text expressions.

To evaluate the significance of $q_i$ on the semantics of $Q_m$, we can evaluate the semantic change before and after removing this word: $S_1(q_i, Q_m, c) = 1 - \cos(c \cup Q_m, c \cup Q_m \setminus \{q_i\}),$
% \small
% \begin{equation}\label{eq:token_relevance}
%     S_1(q_i, Q_m, c) = 1 - \cos(c \cup Q_m, c \cup Q_m \setminus \{q_i\}),
% \end{equation}\normalsize
where $c$ denotes the prompt and
$\cos(\cdot, \cdot)$ is the cosine similarity function. 
% We measure the Cross-Encoder~\citep{reimers2019sentence}-RoBERTa-large~\citep{liu2019roberta} as the similarity since the popular SentenceTransformers Library~\citep{reimers-2019-sentence-bert} can be one of the most powerful sentence similarity evaluation models. 
In real-world applications, larger $S_1(q_i, Q_m, c)$ denotes that removing $q_i$ will lead to significant semantic changing, indicating that $q_i$ is more relevant.

% \noindent \textbf{Uncertainty Proportion.} For the percentage of uncertainty committed by $q_i$, we can simply derive the rate from \eqref{eq:PE}:
% \small
% \begin{equation}\label{eq:token_uncertainty_proportion}
%     \textit{UP}_\textit{T}(q_i, Q_m, c) = \frac{-\log p(q_i | Q_m^{< i}, c)}{\textit{PE}(Q_m, c)},
% \end{equation}\normalsize
% where  larger $\textit{UP}_\textit{T}(q_i, \vs, c)$ denotes that $q_i$ commits more uncertainty if we  estimate the uncertainty of sentence $Q_m$; vice versa. \textit{PE} denotes the effective Predictive Entropy  model as follows:
% \small
% \begin{equation}\label{eq:PE}
%     \textit{PE}(Q_m, c) = - \log p(Q_m|c) = \sum_{i}{-\log p(q_i|Q_m^{< i}, c)}.
% \end{equation}\normalsize
%  \eqref{eq:PE} is interpreted as the accumulation of the word-wise entropy.

\noindent \textbf{Structure-level  augmentation.} Since different users tend to utilize various text structures for video grounding, we need to augment the text structure for more diverse texts. Similarly, we utilize the following prompt for structure-level rewriting: $P_{i,m}\!\leftarrow \texttt{LLM}(P'_{i,m},$ ``Rewrite the text `\textbf{text}' concisely by \underline{changing the text structure} while keeping the meaning'').
% \small
% \begin{equation}
% \begin{split}
%     P_{i,m}\!\leftarrow \texttt{LLM}(P'_{i,m},\text{``Rewrite the text `\textbf{text}' concisely by \underline{changing the text structure} while keeping the meaning''}).
%      \end{split}
% \end{equation}\normalsize
% \textit{``Rewrite the text `\textbf{text}' concisely by changing the text structure while preserving the meaning}''.
% \subsection{Negative text Augmentation}
% \label{subsec:negative}
% In fact, we involve multiple sentences to benefit estimating uncertainty~\citep{kadavath2022language}. For example, \textit{PE} is usually the arithmetic mean of multiple sentences, that is, $\frac{1}{K}\sum_{k}{\textit{PE}(Q_m^k, c)} \, (1 \leq k \leq K)$ where $S=\{Q_m^1, Q_m^2, ..., Q_m^{\textit{K}}\}$ consists of $K$ sentences related to $c$ and $ Q_m^{k} \in S$ is the $k$-th sentence.
Given a sentence $Q_m$, we  define the sentence-level relevance of $Q_m^i$ as the probability-weighted semantic similarity with other sentences: $S_2(Q_m^i,Q_m^j, c) = \sum\nolimits_{j=1,j \neq i}{\cos(Q_m^i, Q_m^j)p(Q_m^j|c)},$ 
% \small
% \begin{equation}\label{eq:sentence_relevance}
%     S_2(Q_m^i,Q_m^j, c) = \sum\nolimits_{j=1,j \neq i}{\cos(Q_m^i, Q_m^j)p(Q_m^j|c)},
% \end{equation}\normalsize
where $p(Q_m^j, c)$ denotes the generative probability that provides more confidence to $Q_m^j$, and higher $p(Q_m^j, c)$ makes $Q_m^j$ more acceptable.
An intuitive observation is that if a sentence is  semantically consistent with other sentences, the sentence is more convincing and more representative. 

Similar to positive text augmentation, we generate negative texts by changing their words and sentence structure. Thus, based on the multi-level language rewriting, we can conduct coarse-grained language alignment for text augmentation.

\subsection{Attribute-based Text Reasoning}
In fact, Section \ref{subsec:positive} only considers the semantics of the sentence itself, ignoring the latent information of the sentence. For example, ``a person is driving a car'' contains two significant objects: ``person'' and ``car''.  ``person'' corresponds to the following attributes: a head, two eyes, two arms, etc, while the attributes of ``car'' include: four wheels, a steering wheel, etc. There attributes will assist VLMs to understand videos and texts for bridging the visual and textual gap.

\noindent \textbf{Attribute generation.}
For some semantically similar sentences, they always have similar attributes. 
% For example, ``person opens the door'' and ``door is opened by person'' have the same attributes. 
Therefore, we generate the attributes for all positive and negative texts. Although embedding attributes can help us to understand the sentence, current VLM models cannot fully understand the latent semantics. For example, ``a person is driving a car'' and ``a car is running on the road'' have similar semantics.
% The pipeline of our method has been presented in Fig.~\ref{fig:method}. As discussed in Sec.~\ref{subsec_preliminary}, the word embedding of a specific class name is concatenated with the learnable tokens for conventional prompt tuning~\citep{Zhou_2022, li2021prefix, lester2021power}.
% % in the final Previous approaches have commonly concatenated soft tokens directly with class names. 
% However, we contend that this practice represents a shortcut for CLIP to attain high accuracy without suitable rationales~\citep{mao2023doubly}. For instance, when presented with a class name of ${\rm bird}$, CLIP may establish a semantic connection with the sky, introducing a dependence on the background rather than capturing the semantics of birds. This reliance on spurious correlations substantially undermines generalization capabilities.
Therefore, rather than directly using the original sentence, we design a model with high confidence in visual attributes. Two intuitions are considered in this model: 1) different from the original sentence, aligning explicitly with visual attributes can push the deigned model to mine the inherent semantics in the given sentence. 2) visual attributes contain more fine-grained features, which can provide more details for cross-modal reasoning.

% \begin{figure}[t!]
% % \vspace{-15pt}
% \centering
% % \subfigbottomskip=-3.6pt
% % \subfigcapskip=-5.5pt
% %\vspace{0.35cm}
%  \includegraphics[width=\linewidth]{filter_01.jpg}
% \vspace{-5mm}
% \caption{
% \small Attribute selection.
% % Jailbreak performance of different categories of malicious questions.  
% }
% \label{fig:attribute}
% \end{figure}

Firstly, we utilize video  and text encoders to extract the video and text features. Since our model is plug-and-play, it does not depend on specific feature encoders. For the fair comparison, we adopt the same video  and text encoders with compared methods. For the text $Q$ with $J$ words, we denote word-level text feature as $f^{W}=\{f^{w}_j\}_{j=1}^{J} \in \mathbb{R}^{J \times d}$ and the sentence-level text feature  as $f^q \in\mathbb{R}^{d}$, where $d$ is feature dimension. Similarly, we denote the extracted video features as $f^{V}=\{f^{v}_i\}_{i=1}^{N_v} \in \mathbb{R}^{N_v \times d}$, where $N_v$ is the frame number.

\begin{figure}
\begin{center}
% \vspace{-7mm}
 \includegraphics[width=\linewidth]{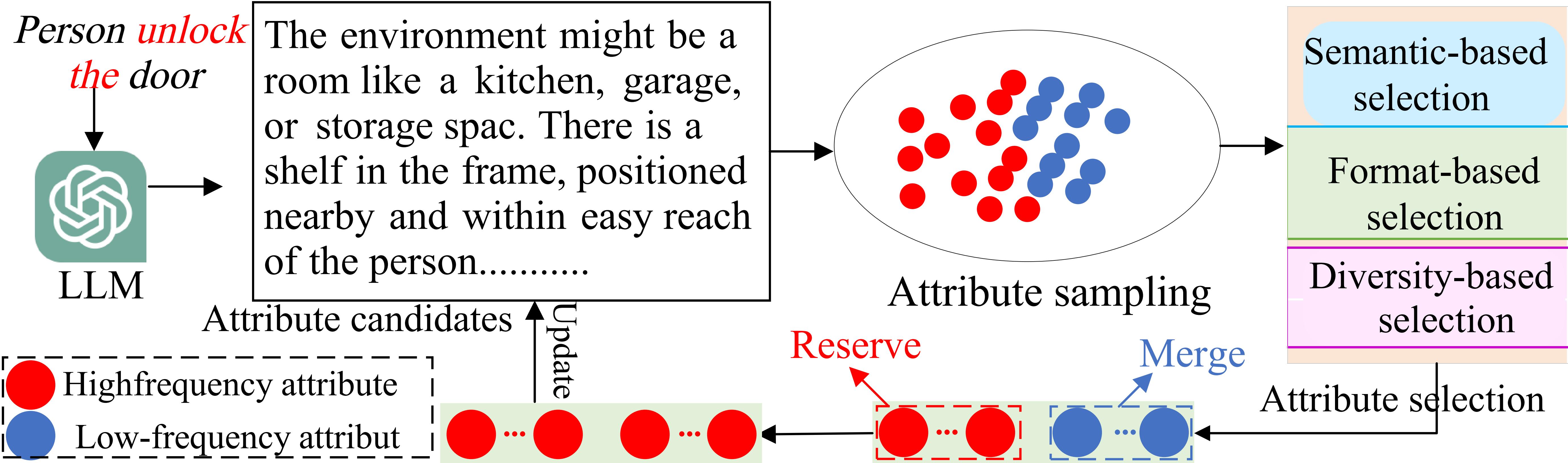}
\end{center}
% \vspace{-3mm}
\caption{
\small Our attribute selection module.
% Jailbreak performance of different categories of malicious questions.  
}
\label{fig:attribute}
% \vspace{-3mm}
\end{figure}

\noindent \textbf{Attribute sampling.}
% \label{sub:attr_samp}
% While LLMs can generate attributes associated with the class names, 
We find that some generated attributes have a stronger semantic correlation with visual features than others, and some attributes have less significance (even may be hallucination information), which will lead to high computational cost. Therefore, removing some low significance can not only decrease the computational cost but also improve the model generalization.
% it's evident that 
% some attributes exhibit a stronger semantic correlation with visual features than others. Our subsequent experiments further highlight that the removal of ineffective attributes not only reduces memory consumption but also improves the model's accuracy.
% actually 
% Moreover, the inclusion of a larger number of attributes in training necessitates correspondingly greater memory resources. 
% Therefore, in this section, we 
As shown in Figure \ref{fig:attribute}, we address the problem by selecting effective attributes from an attribute pool.
% It is essential to note that while our primary task is few-shot adaptation, this method is equally applicable to
% % zero-shot recognition, especially 
% attribute-based zero-shot recognition~\citep{roth2023waffling, pratt2023does, menon2022visual}.
Two main criteria are utilized during the attribute selection: Firstly, we prioritize attributes that are both representative and non-redundant. Secondly, we seek attributes with the highest semantic relevance to the images when compared to other attributes.
% Our selection process revolves two main criteria: 1) the selected attributes should be
% % Firstly, we prioritize attributes that are 
% both representative and non-redundant; 2) the selected attributes should be semantically related to the class-specific images.
% % . Secondly, we seek attributes with the highest semantic relevance to the images when compared to other attributes. 
Finally, we use the following steps for attributes: 1) For the attributes $a_m$ associated with sentence $Q$, we partition them into $N_c$ clusters based on their feature similarity. This clustering strategy aims to ensure that each cluster represents a distinct aspect, \textit{e.g.}, color or shape, in the descriptions. 2)  In each cluster, we rank the attributes by assessing their similarity to visual features, and select the one with the highest relevance. By the above strategy, the following attributes will be filtered out: non-visual attributes  and  incorrect visual attributes that are semantics-unrelated to the videos. To obtain optimal attributes, we introduce three  attribute selection strategies:

% An illustrative example could be found in ImageNet-Sketch~\citep{wang2019learning}, where the predominant content comprises sketches, devoid of the real colors of objects. Nevertheless, LLMs tend to generate class-specific colors despite careful prompting, e.g., $\rm red$ for $\rm apple$. In this situation, our attribute sampling approach initially groups attributes related to color into one cluster and subsequently identifies the most pertinent colors for sketches, i.e., ${\rm black \ and \ white}$. Fig. ~\ref{fig:attr_samp} offers concrete examples of this process.

% \noindent \textbf{Attribute filtering.}
% % \label{sec:filtering}
% Although we can obtain many attributes during attribution generation, there will be some noises in these attributes.  For example, a text ``person visit the shop sometime'' and its generated attribute ``person dislikes the long flight'' have  no logical relevance. Thus, we need to filter out the proper text-attribute pairs from the attribute pool for better attribute quality.

\noindent \textbf{Semantic-based  selection.}
Firstly, we want to make the sentence text has the similar semantics with its generated attributes. Since the natural language inference (NLI) model \cite{chen2017enhanced} can mine the relationship between texts and the attributes by inferring the logical entailment, we introduce an NLI-based binary filter  ($f_{nli}$) as a critic, and discard the pairs which do not achieve the entailment score over the threshold $\gamma_{1}$: $O_{1}(x, y) = \mathbb{1} \{  f_{nli}(x \Rightarrow y) \ge \gamma_{1} \},$
% \small
% \begin{equation}
% % \small
% \begin{split}
%     O_{1}(x, y) = \mathbb{1} \{  f_{nli}(x \Rightarrow y) \ge \gamma_{1} \},
% \end{split}
% \end{equation}\normalsize
where $x$ denotes the input, and $y$ means the output.

% A faithful paraphrase should preserve the semantics of the source statement without hallucinating unsupported content. NLI models are well-suited to quantify this relationship, as they are trained to infer the logical entailment between an arbitrary pair of statements \citep{nli-verifier}. Hence, we define a binary filter using a small NLI model \citep{wanli} as a critic, and discard the pairs that do not achieve the entailment score over the threshold $\tau_{\textit{semantic}}$:
% \begin{equation*}
% \small
% \begin{split}
%     f_{\textit{semantic}}(x, y) = \mathbbm{1} L\{ & p_{\textit{NLI}}(x \Rightarrow y) \ge \tau_{\textit{semantic}} \,\, \land \\
%                                                     & p_{\textit{NLI}}(y \Rightarrow x)) \ge \tau_{\textit{semantic}} R\}
% \end{split}
% \end{equation*}

\noindent \textbf{Format-based  selection.}
When we rewrite the given sentence, we need to make the format of the given sentence various, and preserve its original meaning. Thus, we want to filter the origin-variance pair to learn the format-free dissimilarity. Especially, two metrics are used to evaluate the dissimilarity: 1) the token overlap between different sentences and 2) their syntactic difference. For the first, we filter the pairs with a higher Rouge-L~\citep{Lin2004ROUGEAP} than a threshold $\gamma_{3}$. As for the syntactic difference, we first parse the constituency tree of the origin and variety, and then filter the pairs based on their tree edit distance: $O_2(x, y) = \mathbb{1} \{  D_t(x, y) \ge \gamma_{2} \,\, \land 
                                                         f_{rou}(x, y) \le \gamma_{3}\},$
% \small
% \begin{equation} \label{f_dissim}
% % \small
% \begin{split}
%     O_2(x, y) = \mathbb{1} \{  D_t(x, y) \ge \gamma_{2} \,\, \land 
%                                                          f_{rou}(x, y) \le \gamma_{3}\},
% \end{split}
% \end{equation}\normalsize
where $D_t(\cdot, \cdot)$ denotes the tree edit distance. The two dimensions of dissimilarity complement each other.  On the one hand,  $f_{rou}(\cdot, \cdot)$ promotes lexical divergence in each pair. On the other hand, $D_t(\cdot, \cdot)$ can be used to preempt ``hacking'' the word-overlap metric by  simply switching a few words in the source sentence with corresponding synonyms.

\noindent \textbf{Diversity-based  selection.}
For  sentence rewriting, we need a diverse range of generated sentences since the diversity of attributes can directly affect the robustness of the trained model. Therefore, we introduce a critic $O_3$ for the diversity. We define two pairs $(x_1, y_1)$ and $(x_2, y_2)$ to be duplicates when one pair entails another, either on the input side ($x_1 \Rightarrow x_2$) or on the output side ($y_1 \Rightarrow y_2$). In the diversity filter, we first cluster all entailing pairs, and then discard all but one with the largest entailment score. Thus, we can utilize the graph traversal for the diversity filter. 

Based on the above critics, we can filter the attribute candidate pool $\mathcal{A}$ into an updated pool $\mathcal{U}$: $    \mathcal{U} = \{(x, y)|  (x, y) \in \mathcal{A}, 
     O_1 \land O_2 \land O_3(x,y) = 1\}.$
% \small
% \begin{equation}
% \begin{split}
%     \mathcal{U} = \{(x, y)|  (x, y) \in \mathcal{A}, 
%      O_1 \land O_2 \land O_3(x,y) = 1\}.
% \end{split}
% \end{equation}\normalsize

\subsection{Video-guided Self-weighted Cross-modal  Bridging}
% \subsection{Video-guided Positive-negative Text  Incorporation}
\label{subsec:weight_module}
% \subsection{Weighted Sentence  Incorporation for Cross-modal Fusion}
% \label{subsec:weight_module}
In fact, different words (\textit{e.g.}, noun, verb, and adjective) have distinct significance in text understanding. For instance, some adjectives are more important for video grounding in some cases, while some verbs are more significant for distinguishing different target moments.  Previous VLM methods treat all sentence components equally, which might limit these methods to fully understand the entire sentence. For example, if there is no adjective in the anchor text, the negative text with adjectives cannot contribute to our model since the adjective is not discriminative for the text.
Thus, we aim to analyze the relative significance of each word to adaptively integrate different words, where we adaptively predict the salience of sentence components for each anchor text. Without any supervision, we can obtain the significance score which means which word is more significant for text understanding.
Therefore, we can find an optimal integration strategy of sentence components, which makes VLM selectively understand different sentence components for a query.

\noindent \textbf{Cross-modal bridging.}
Based on these positive and hard negative samples, we can encourage the designed VLM models to distinguish the difference between different words in each sentence part.
% Besides, by these generated positive and negative  samples, we can better explore their relevance with the video, which enhances the video-text interaction for the VLM task. For supervising the VLM model to understand the 
For supervising the VLM model to understand the text input, we introduce the following loss based on three types of text input:
\scriptsize
\begin{equation}\label{L_contrast}
% \small
    \mathcal{L}_{cl}^i\! =\! -\! \log \frac{\beta \cdot  \exp[1/ \tau \cdot \cos{(f^{V}\!,\! g_i^{n})}]}{(\!1\!-\!\beta\!)\! \cdot\exp[1/ \tau \cdot \cos{(f^{V}\!,\! g_i^{p_i})}]\! +\! \beta \! \cdot \!  \exp[1\!/\! \tau\! \cdot\! \cos\!{(\!f^{V}\!,\! g_i^{n}\!)}\!]}\!,\nonumber
\end{equation}\normalsize
where $\beta\in(0,1)$ is a parameter;
$g_i$ denotes the $i$-th text; $ g_i^{n_{i,j}}$ and $g_i^{p_i}$ denote the the negative text and the positive text, respectively; $\tau$ denotes the temperature parameter. 
% $\mathbf{S}_i$ denotes the set of all negative texts of $i$-th text. 
By  $\mathcal{L}_{cl}^i$, we can enhance the effectiveness of the designed model by these generated auxiliary texts.

\begin{table}[t!]
% \vspace{-8mm}
% \small
% \caption{\small Text-to-video and video-to-text retrieval comparisons on MSR-VTT.
% } 
 \scriptsize
\setlength\tabcolsep{0.15cm}
\begin{center}
% \scalebox{0.75}{
% \setlength{\tabcolsep}{0.2mm}{
\begin{tabular}{c|ccc|ccccccc} 
    \hline 
     \multirow{2}{*}{Method} & \multicolumn{3}{c|}{Without text augmentation  } & \multicolumn{3}{c}{With text augmentation  } \\
     \cline{2-7}
     & R@1{$\uparrow$} & R@5{$\uparrow$} & R@10{$\uparrow$} &  R@1{$\uparrow$} & R@5{$\uparrow$} & R@10{$\uparrow$}  \\
    \hline
    \multicolumn{7}{c}{Text-to-video retrieval}\\
    \hline
    {{\textit{{CLIP-ViT-B/32}}}} & & & & & & \\ 
    X-Pool  & 46.9 & 72.8 & 82.2  & 40.1& 68.2& 76.5 \\
    \textbf{+Ours } &  \textbf{47.8}& \textbf{74.9}& \textbf{83.5}&  \textbf{45.9}& \textbf{72.3}& \textbf{81.4}  \\\hline
         % UATVR~\citep{fang2023uatvr}  & 47.5 & 73.9 & 83.5 & 2.0 & 12.3  & -- & -- & -- & -- & -- \\
    % DiffusionRet~\citep{jin2023diffusionret}   & 49.0 & 75.2 & 82.7 & 2.0 & 12.1 & 24.4 & 43.1 & 54.3 & 8.0 & {40.7} \\
    % \textbf{+Ours } & \textbf{51.7}& \textbf{77.9}& \textbf{84.5}& \textbf{1.0}& \textbf{11.8}& \textbf{25.7}& \textbf{45.2}& \textbf{55.8}& \textbf{7.0}& \textbf{38.5} \\\hline
    % TEFAL~\citep{ibrahimi2023audio} &49.4 & {75.9} & \textbf{83.9}& 2.0& 12.0& 26.8 &46.1 &56.5 &7.0 &44.4 \\
    % \textbf{+Ours } & \textbf{51.9}& \textbf{77.4}& 83.5& \textbf{1.0}& \textbf{11.8}& \textbf{28.4}& \textbf{46.9}& \textbf{58.2}& \textbf{6.0}& \textbf{42.1} \\\hline
    CLIP-ViP & 50.1 & 74.8 & 84.6 & 42.3& 69.4& 77.8 \\
    \textbf{+Ours } & \textbf{51.7}& \textbf{75.3}& \textbf{85.9}&  \textbf{50.4}& \textbf{73.6}& \textbf{84.5} \\\hline
    T-MASS   &{50.2} &75.3
    &{85.1}  & 42.1& 68.9& 79.2\\
    \textbf{+Ours } & \textbf{52.3}& \textbf{77.9}& \textbf{87.6}& \textbf{51.4}& \textbf{70.2}& \textbf{81.3}  \\
    \hline \hline 
    {\textit{{CLIP-ViT-B/16}}}  & & & & & & \\
    X-Pool &48.2 &73.7 &82.6  & 39.7& 68.5& 78.4 \\
    \textbf{+Ours } &  \textbf{50.7}& \textbf{76.2}& \textbf{85.2}& \textbf{48.9}& \textbf{75.3}& \textbf{84.0} \\\hline
    % UATVR~\citep{fang2023uatvr} & 50.8 & 76.3 & 85.5 & 1.0 & 12.4  & -- &-- &-- & --&-- \\ 
    CLIP-ViP & {54.2} & {77.2} & 84.8  & 51.2& 73.9& 80.4  \\
    \textbf{+Ours } & \textbf{56.8}& \textbf{79.4}& \textbf{85.9} & \textbf{53.6}& \textbf{77.8}& \textbf{84.2} \\\hline
    T-MASS   &52.7 &77.1 &{85.6}  & 49.2& 70.5& 83.9  \\
    \textbf{+Ours } & \textbf{54.9}& \textbf{82.6}& \textbf{86.8}& \textbf{53.4}& \textbf{81.0}&\textbf{ 86.2}\\\hline
    \hline
        \multicolumn{7}{c}{Video-to-text retrieval}\\
    \hline
     {\textit{CLIP-ViT-B/32}} & & & & & \\
    % CLIP4Clip~\citep{luo2022clip4clip} & 42.7 & 70.9 & 80.6 & 2.0 & 11.6   \\
    % \textbf{+Ours } & \textbf{44.2}& \textbf{73.8}& \textbf{84.3}&  \textbf{2.0}& \textbf{10.4} \\ \hline 
    % CenterCLIP~\citep{zhao2022centerclip}& 42.8 &71.7 &82.2& 2.0 &10.9 \\
    %  \textbf{+Ours } & \textbf{44.5}& \textbf{73.0}& \textbf{84.1}& \textbf{1.0}& \textbf{9.7} \\ \hline 
    X-Pool & 44.4 & 73.3 & 84.0  & 41.2& 68.5& 80.4 \\
    \textbf{+Ours } & \textbf{45.8}& \textbf{76.4}& \textbf{87.3} & \textbf{42.7}& \textbf{74.5}& \textbf{86.0} \\ \hline 
    % TS2-Net~\citep{liu2022ts2} & 45.3 & 74.1 & 83.7 & 2.0 & 9.2 \\ 
    % \textbf{+Ours } & \textbf{48.6}& \textbf{77.5}& \textbf{85.7}& \textbf{2.0}& \textbf{7.9}\\ \hline 
    % DiffusionRet~\citep{jin2023diffusionret} & 47.7 & 73.8 & 84.5 & 2.0 & 8.8  \\
    % \textbf{+Ours } & \textbf{51.0}& \textbf{75.9}& \textbf{87.4}& \textbf{2.0}& \textbf{6.9} \\ \hline 
    UATVR & 46.9 & 73.8 & 83.8  & 43.0& 67.9& 78.3 \\
    \textbf{+Ours } &  \textbf{49.7}& \textbf{75.6}& \textbf{86.4} & \textbf{47.8}& \textbf{74.0}&\textbf{ 83.9} \\ \hline 
    T-MASS   &{47.7} &{78.0} &{86.3} & 42.9& 73.5& 82.6 \\ 
    \textbf{+Ours } & \textbf{ 51.5}& \textbf{79.9}& \textbf{89.8} & \textbf{49.5}& \textbf{78.1}& \textbf{87.5} \\ \hline 
    \hline 
    {\textit{CLIP-ViT-B/16}} & & & & & \\
    X-Pool & 46.4 & 73.9  & 84.1 & 42.8& 72.0& 81.6   \\
    \textbf{+Ours } & \textbf{50.2}& \textbf{77.4}& \textbf{86.3} & \textbf{48.7}& \textbf{76.0}& \textbf{84.2} \\ \hline 
    % TS2-Net~\citep{liu2022ts2}  &46.6 & 75.9 & 84.9 & 2.0 & 8.9 \\ 
    %  \textbf{+Ours } &  \textbf{48.8}& \textbf{78.3}& \textbf{86.1}& \textbf{1.0}& \textbf{7.6} \\ \hline 
    % CenterCLIP~\citep{zhao2022centerclip}& 47.7 &75.0& 83.3& 2.0 &10.2\\
    % \textbf{+Ours } &\textbf{ 49.8}& \textbf{78.0}&\textbf{ 86.4}& \textbf{2.0}& \textbf{6.5} \\ \hline 
    UATVR & 48.1& 76.3&85.4  & 41.6& 73.0& 81.9 \\
     \textbf{+Ours } & \textbf{50.9}& \textbf{77.4}& \textbf{90.5}& \textbf{48.9}& \textbf{76.3}& \textbf{87.9} \\ \hline
    T-MASS  &{50.9} &{80.2} &{88.0}  & 48.3& 75.6& 84.9\\ 
     \textbf{+Ours } & \textbf{53.7}& \textbf{84.2}& \textbf{91.5}&  \textbf{50.8}& \textbf{82.7}& \textbf{90.4}\\ \hline
\end{tabular}
% }}
\end{center}
% \vspace{-5mm}
\caption{ Video text retrieval comparisons on MSR-VTT.
} 
% \vspace{-3mm}
\label{tab:videoqa}
% \vspace{-20mm}
\end{table}

\begin{table}[t!]
\centering
% \vspace{-8mm}
% \resizebox{0.5\textwidth}{!}{
 \scriptsize
\setlength\tabcolsep{0.12cm}
% \scalebox{0.85}{
% \setlength{\tabcolsep}{0.9mm}{
\begin{tabular}{c|cccc|cccccccc}
\hline
 \multirow{2}*{Method(\# Frames)} &\multicolumn{4}{|c|}{w/o text augmentation}& \multicolumn{4}{c}{w/ text augmentation}\\\cline{2-9}
~& Int\small{$\uparrow$} & Seq\small{$\uparrow$} & Pre\small{$\uparrow$} & Fea\small{$\uparrow$} & Int\small{$\uparrow$} & Seq\small{$\uparrow$} & Pre\small{$\uparrow$} & Fea\small{$\uparrow$} \\
\hline 
All-in-One (32)   & 47.5 & 50.8 & 47.7 &  44.0 & 42.9& 48.5& 44.0& 40.2 \\
\textbf{+Ours (32)} & \textbf{48.3}& \textbf{51.9}& \textbf{49.6}& \textbf{45.7} & \textbf{47.9}& \textbf{51.3}& \textbf{48.7}& \textbf{44.3} \\\hline
% MIST (32)   &  55.5 &  54.2 & 54.2  & 44.4 & 50.7& 51.4& 50.2& 38.4  \\
% \textbf{+Ours (32)} &\textbf{ 58.6}& \textbf{59.5}& \textbf{58.4}& \textbf{47.0} & \textbf{57.0}& \textbf{56.3}& \textbf{57.2}& \textbf{45.8} \\\hline
InternVideo (8)   &   62.7 &  65.6 &  54.9  &  51.9 &  55.6& 61.0& 50.3& 47.2 \\
\textbf{+Ours (8)} &\textbf{ 63.8}& \textbf{67.7}& \textbf{58.9}& \textbf{55.2} & \textbf{61.8}& \textbf{64.9}&\textbf{ 57.4}& \textbf{54.3} \\\hline
% SeViLA (4)   &  63.7 & {70.4} &  {63.1}  &  {62.4} & 58.7& 62.2& 57.9& 57.8 \\
% \textbf{+Ours (4)} & \textbf{66.7} & \textbf{72.9}& \textbf{66.4}& \textbf{65.3} & \textbf{65.3}& \textbf{68.9}& \textbf{64.}2& \textbf{63.}4 \\\hline
BLIP-2 (4)   &  {65.4}  &  69.0  &   59.7  &  54.2 & 60.9& 66.3& 54.3& 50.1  \\
\textbf{+Ours (4)} & \textbf{67.8}& \textbf{72.5}& \textbf{61.4}& \textbf{56.8} & \textbf{66.2}& \textbf{71.6}& \textbf{58.7}& \textbf{55.3}\\\hline
% \textbf{ViLA (2) (Ours)} &   65.0 & 65.4 & 62.2  & 58.8 &  62.9 & 72\\
%BLVQA (4) (Ours)  &   69.5 & 70.4 & 63.9  & 60.6 &  \textbf{66.1} \\
%ViLA (4) (Ours)  &   69.3 & 70.0 & 63.9  & 64.3 &  \textbf{66.5} \\
%ViLA (8) (Ours)  &   70.6 & 74.1 & 66.3  & 60.6 &  \textbf{67.9} \\
\end{tabular}
% }}
% \vspace{-3mm}
\caption{{ Comparison Results on STAR VideoQA.} 
% For Interaction type of question, our ViLA improved \textbf{4.6\%}.
% On average, our ViLA out-performs the SOTA method by 2.2\% when using 4 frames with {\bf3.04$\times$} speed up.
%set- ting and achieved new SOTA at 67.9\% on this dataset. 
% We \textbf{bold} the best results and \underline{underline} the second best.
% Note that our ViLA using 2-frames out-performs BLIP-2 using 4-frames.
%We achieved state-of-the-art fro all different types of questions and especially for Interaction.
}
\label{tab:star}
% \vspace{-6mm}
\end{table}

\begin{table*}[t]
 % \vspace{-4mm}
 % \small
\centering
 % \resizebox{0.5\textwidth}{!}{
% \scalebox{0.7}{
 \scriptsize
\setlength\tabcolsep{0.24cm}
\begin{tabular}{c|ccc|ccc||c|ccc|ccccccc}
\hline
\multirow{2}*{Method(\# Frames)} & \multicolumn{3}{|c|}{w/o text augmentation}& \multicolumn{3}{c||}{w text augmentation} &\multirow{2}*{Method(\# Frames)} & \multicolumn{3}{|c|}{w/o text augmentation}& \multicolumn{3}{c}{w text augmentation} \\\cline{2-7} \cline{9-14}
~  & Tem\small{$\uparrow$} & Cau\small{$\uparrow$} & Des\small{$\uparrow$} & Tem\small{$\uparrow$} & Cau\small{$\uparrow$} & Des\small{$\uparrow$} &~  & Tem\small{$\uparrow$} & Cau\small{$\uparrow$} & Des\small{$\uparrow$} & Tem\small{$\uparrow$} & Cau\small{$\uparrow$} & Des\small{$\uparrow$} \\
\hline 
All-in-One(32)     & 48.6 & 48.0 & 63.2 & 40.2& 37.9& 53.8 &InternVideo(8)  &  58.5 & 62.5 &  75.8 & 52.9& 57.4& 70.3 \\
\textbf{+Ours(32)} & \textbf{50.1}& \textbf{51.9}& \textbf{64.7} & \textbf{48.6}& \textbf{50.2}& \textbf{61.3} &\textbf{+Ours(8)}  & \textbf{62.5} & \textbf{66.3}& \textbf{76.4} & \textbf{61.8}& \textbf{59.7}& \textbf{74.5} \\\hline 
Just Ask(20)   & 51.4 & 49.6 & 63.1 & 42.7& 40.1& 54.0 &BLIP-2(4)  & 67.2  & 70.3 &  79.8 & 64.0& 61.9& 72.3 \\
\textbf{+Ours(20)} & \textbf{54.3}& \textbf{52.9}& \textbf{67.8} & \textbf{50.9}& \textbf{49.3}& \textbf{62.7 } &\textbf{+Ours(4)}  & \textbf{70.1}& \textbf{72.9}& \textbf{80.4 } & \textbf{69.2}& \textbf{70.1}& \textbf{78.4}\\\hline 
MIST(32)     &  56.6 & 54.6 &  66.9 & 51.9& 48.2& 55.3 &SeViLA(4)   &  {{67.7}}  & {{72.1}} & {{82.2}} & 64.0& 66.8& 76.9\\
\textbf{+Ours(32)} & \textbf{60.3}& \textbf{56.9}&\textbf{ 69.8}& \textbf{57.2}& \textbf{55.4}& \textbf{67.9} & \textbf{+Ours(4)}  & \textbf{72.4}& \textbf{74.9}& \textbf{85.3} & \textbf{70.5}& \textbf{72.7}& \textbf{83.9}\\\hline
HiTeA(16)    & 58.3  & 62.4 &  75.6 & 52.2& 57.6& 59.3 & VideoQA-TA(4)& 58.9& 63.2& 76.0& 52.1& 57.9& 58.5\\
\textbf{+Ours(16)}  & \textbf{62.8}& \textbf{65.7}& \textbf{77.3} & \textbf{60.4}& \textbf{63.9}& \textbf{74.9} & \textbf{+Ours(4)}& 63.4& 66.1& 77.9& 61.8& 64.7& 76.5\\\hline
% InternVideo(8)  &  58.5 & 62.5 &  75.8 & 52.9& 57.4& 70.3 \\
% \textbf{+Ours(8)}  & \textbf{62.5} & \textbf{66.3}& \textbf{76.4} & \textbf{61.8}& \textbf{59.7}& \textbf{74.5} \\\hline
% BLIP-2(4)  & 67.2  & 70.3 &  79.8 & 64.0& 61.9& 72.3 \\
% \textbf{+Ours(4)}  & \textbf{70.1}& \textbf{72.9}& \textbf{80.4 } & \textbf{69.2}& \textbf{70.1}& \textbf{78.4} \\\hline
% SeViLA(4)   &  {{67.7}}  & {{72.1}} & {{82.2}} & 64.0& 66.8& 76.9 \\
% \textbf{+Ours(4)}  & \textbf{72.4}& \textbf{74.9}& \textbf{85.3} & \textbf{70.5}& \textbf{72.7}& \textbf{83.9} \\\hline
% VideoQA-TA(4)& 58.9& 63.2& 76.0& 52.1& 57.9& 58.5 \\
% \textbf{+Ours(4)}& 63.4& 66.1& 77.9& 61.8& 64.7& 76.5\\\hline
% SeViLA~\citep{yu2023self}    & 8 &  {{67.0}}  & {{73.8}} &  {{81.8}} \\
% \textbf{+Ours} & \textbf{8} & \textbf{72.8}& \textbf{76.9}& \textbf{87.2} \\\hline
% \midrule
%ViLA (4) (Ours)  &   69.1 & 73.6 & 81.3  &  \textbf{74.2} \\
% \textbf{ViLA (Ours)}  & 4 (8 to 4) &   {\textbf{71.0}} & 72.9 & {\textbf{82.7}}  \\
% \textbf{ViLA (Ours)}  & 4 (32 to 4) &   70.1 & {\textbf{73.8}} & 82.1  \\
%ViLA (Ours)  & 16 &  69.5 & 74.0 &  81.7 &  \textbf{75.0} & 188M \\
% \midrule
% %ViLA (Ours)  & 32 &  \textbf{72.3}  & \textbf{74.9} &  82.1 &  \textbf{75.6} & 188M & -\\
% \textbf{ViLA+LoRA (Ours)}  & \textbf{4} &  71.4  & 74.5 &  80.3 \\
\end{tabular}
% }
% }
% \vspace{-3mm}
\caption{ VideoQA performance comparison on NExT-QA, where the value means the accuracy of providing the right answer. 
% Here we measure the accuracy of choosing the right answer. 
%Our ViLA outperforms the SOTA method by {\bf 0.9-1.0\%} at 4 frames with {\bf1.45-3.04$\times$} speed up and  by {\bf1.0\%} at 8 frames with {\bf1.35$\times$} speed up.
%and push the accuracy to reach 75.5\% on this dataset at 32 frames setting. 
%And our proposed ViLA with 4 frames even achieved better performance than SeViLA with 8 frames. 
% Especially on Temporal and Causal type of questions, our ViLA (using only 4 frames) improves {\bf 3.3\%} and {\bf 1.7\%} respectively, compared with SeViLA. 
% We use \textbf{bold-face} font to indicate the best results and \underline{underline} on the second best using the same number of frames.
% ViLA using 2-frames only out-performs BLIP-2 using 4-frames by \textbf{1.3\%}.  ViLA also achieves upto {\bf 3.04$\times$} speedup. It needs to be noted that our ViLA achieves 75.1\% average accuracy with only 4 frames when we finetune LLM with LoRA~\citep{hu2021lora}.
%The comparison with SeViLA shows the effectiveness of our Frame-Prompter and QFormer-Distillation.
%\ljb{can you have the brown box for the text 'brown box' and the same for blue box as well?}
}
\label{tab:nextqa}
 % \vspace{-4mm}
\end{table*}

\begin{table*}[t!]
% \small
 \scriptsize
\setlength\tabcolsep{0.07cm}
 % \vspace{-0.08cm}
    \centering
    % , where ``MomentDiff*'' denotes ``we report the results of MomentDiff on ActivityNet-CD''
% \vspace{-3mm}
    % \scalebox{0.85}{
    % \setlength{\tabcolsep}{0.4mm}
    \begin{tabular}{c|c|cccc|cccc|cccc|ccccccc}
    \hline
    \multirow{4}*{Method} & \multirow{4}*{Type} &\multicolumn{8}{c|}{ActivityNet Captions} & \multicolumn{8}{c}{Charades-STA}\\\cline{3-18}
    && \multicolumn{4}{c|}{w/o text augmentation} & \multicolumn{4}{c|}{w/ text augmentation}& \multicolumn{4}{c|}{w/o text augmentation} & \multicolumn{4}{c}{w/ text augmentation}\\\cline{3-18}
    % \cline{3-10}
    ~&~& R@1, & R@1, & R@5, & R@5, & R@1, & R@1, & R@5, & R@5, & R@1, & R@1, & R@5, & R@5, & R@1, & R@1, & R@5, & R@5, \\ 
    ~ & ~ & IoU=0.3 & IoU=0.5 & IoU=0.3 & IoU=0.5& IoU=0.3 & IoU=0.5 & IoU=0.3 & IoU=0.5& IoU=0.5 & IoU=0.7 & IoU=0.5 & IoU=0.7& IoU=0.5& IoU=0.7 & IoU=0.5 & IoU=0.7\\ \hline 
    % \multicolumn{10}{c}{ActivityNet Captions}\\ \hline
% VSA-RNN & FS & 39.28 & 24.43 & 70.84 & 55.52&VSA-RNN & FS & 10.50 & 4.32 & 48.43 & 20.21 \\
% VSA-STV & FS & 41.71 & 24.01 & 71.05 & 56.62& VSA-STV & FS & 16.91 & 5.81 & 53.89 & 23.58 \\
%  MomentDiff &FS& 43.54& 25.72& 69.55&63.72& MomentDiff &FS& 26.81& 20.14& 50.95& 38.23\\
% CTRL & FS & 20.27& 19.40& 47.82& 40.78  \\
% \textbf{+Ours }&\textbf{ FS}& \textbf{22.85}& \textbf{21.73}& \textbf{48.66}& \textbf{43.12}\\\hline
%TGN & FS & 43.81 & 27.93 & 54.56 & 44.20 \\
%LGI \citep{mun2020local}&FS&58.52&41.51&-&-\\
2D-TAN & FS & 59.45 & 44.51 & 85.53 & 77.13& 48.32& 29.38& 71.36& 62.30 & 39.81& 23.25& 79.33& 52.15  &20.18& 11.35&47.05&33.82 \\
\textbf{+Ours}& \textbf{FS}& \textbf{60.46}&\textbf{ 45.29}&\textbf{ 87.94}& \textbf{77.43}&
\textbf{51.86}& \textbf{32.64}& \textbf{72.98}& \textbf{63.75}& \textbf{40.27}& \textbf{24.95}& \textbf{82.96}& \textbf{53.28} & \textbf{23.99}& \textbf{14.75}& \textbf{49.22}& \textbf{34.18}\\\hline
%  DRN & FS & 48.94& 30.26& 69.34& 64.79   \\
%  \textbf{+Ours}& \textbf{FS}& \textbf{50.75}& \textbf{32.92}& \textbf{73.86}& \textbf{67.48}\\\hline
% %FVSG&FS&60.63&45.05&86.11&77.42\\
% RaNet &FS& 52.93& 31.42& 73.80& 65.73 \\
% \textbf{+Ours}& \textbf{FS}& \textbf{53.88}& \textbf{33.74}& \textbf{75.96}& \textbf{68.31}\\\hline
 % MIGCN & FS& 53.76& 30.15& 71.35& 66.27\\
% IVG-DCL & FS & 63.22 & 43.84 & - & - \\
MMN &FS&65.05&48.59&87.25&79.50& 55.30& 31.76& 74.88& 71.52 & 47.31& 27.28& 83.74& 58.41 & 25.33&18.80& 45.97& 35.08\\
\textbf{+Ours}& \textbf{FS}& \textbf{66.05}& \textbf{49.31}&\textbf{ 89.75}& \textbf{81.27}&
\textbf{58.76}& \textbf{33.08}& \textbf{75.33}& \textbf{73.59}& \textbf{49.07}& \textbf{29.32}& \textbf{85.06}& \textbf{60.13} & \textbf{26.87}& \textbf{22.48}& \textbf{46.03}& \textbf{37.85}\\\hline
G2L  &FS & -&51.68& -& 81.32& 55.75& 33.01& 75.25& 70.89  & 47.91& 28.42& 84.80& 59.33  & 26.54& 19.85& 48.06& 36.70   \\
\textbf{+Ours} & \textbf{FS} &  \textbf{66.34} & \textbf{54.26} & \textbf{91.77} & \textbf{84.29} & \textbf{60.90}&  \textbf{46.86}& \textbf{84.39}&\textbf{ 80.62} & \textbf{55.77}& \textbf{32.97} & \textbf{91.38} & \textbf{60.39}  &  \textbf{34.85}& \textbf{27.96}& \textbf{74.28}& \textbf{46.70} \\
\hline
BM-DETR&FS& -& 49.62&-&-&56.24& 33.52& 75.34& 70.26&54.42& 33.84& -&-&28.10& 21.43& 47.26& 35.29 \\
\textbf{+Ours} & \textbf{FS} & \textbf{66.85} & \textbf{54.35} & \textbf{92.13} & \textbf{85.13} & \textbf{61.98}& \textbf{47.93} & \textbf{85.27}& \textbf{81.38}& \textbf{56.20} & \textbf{33.51}& \textbf{92.15}& \textbf{60.92}& \textbf{35.75}& \textbf{28.74}& \textbf{75.13}& \textbf{47.82} \\
\hline
%CTF & WS & 44.30 & 23.60 & - & -\\
% ICVC & WS & 21.88& 18.59& 42.76 & 36.82 \\
% \textbf{+Ours} & \textbf{WS} & \textbf{22.36}& \textbf{20.17}& \textbf{44.19}& \textbf{38.77}\\\hline
% %MARN & WS & 47.01 & 29.95 & 72.02 & 57.49\\
% %SCN & WS & 47.23 & 29.22 & 71.45 & 55.69\\
% LCNet  & WS & 30.15& \textbf{22.08}& 45.80& 39.25  \\
% \textbf{+Ours} & \textbf{WS} & \textbf{33.94}& 21.73& \textbf{46.35}& \textbf{40.57}\\\hline
%CCL & WS & 50.12 & 31.07 & 77.36 & 61.29 \\
VCA & WS & 50.45 & 31.00 & 71.79 & 53.83 & 31.74& 25.37& 46.98& 42.76 &38.13& 19.57& 78.75& 37.75&  17.87& 12.39& 45.70& 22.13\\
\textbf{+Ours} & \textbf{WS} &  \textbf{51.72}& \textbf{33.19}& \textbf{72.85}& \textbf{55.11}&  \textbf{32.99}& \textbf{28.56}& \textbf{48.31}& \textbf{44.07} & \textbf{40.95}& \textbf{20.31}& \textbf{80.42}& \textbf{39.26}  & \textbf{18.63}& \textbf{15.72}& \textbf{46.17}& \textbf{23.88}\\\hline
WSTAN & WS & 52.45 & 30.01 & 79.38 & 63.42&33.72& 25.74& 49.30& 45.88 &29.35& 12.28& 76.13& 41.53 &  8.15 & 5.43& 35.27& 11.86\\
\textbf{+Ours} & \textbf{WS} & \textbf{53.10}&\textbf{ 31.56}& \textbf{80.24}& \textbf{65.77} & \textbf{35.20}& \textbf{27.99}& \textbf{51.84}& \textbf{48.69} & \textbf{30.24}& \textbf{14.06}& \textbf{77.35}& \textbf{42.99}  & \textbf{10.77}& \textbf{6.92}& \textbf{37.40}& \textbf{13.88}\\\hline
%CRM & WS & 55.26 & 32.19 & - & - \\
CNM &WS&55.68&33.33&-&-& 35.72& 28.95& 50.06& 48.72 & 35.15& 14.95& -& -&  14.34& 9.65& 43.88& 18.79\\
\textbf{+Ours} & \textbf{WS} &\textbf{ 56.11}& \textbf{34.08}& \textbf{81.09}& \textbf{67.34}& \textbf{39.56}& \textbf{31.77}& \textbf{52.88}& \textbf{51.99} & \textbf{35.72}& \textbf{16.33}& \textbf{76.52}& \textbf{43.18} & \textbf{16.83}& \textbf{12.05}& \textbf{45.60}& \textbf{21.64}\\\hline
MCMT&WS&58.82& 33.97&-&-& 36.21& 29.52& 50.78& 48.21& 36.23& 15.29& -& -& 15.75& 9.84& 44.21& 19.67\\
\textbf{+Ours} & \textbf{WS} & \textbf{56.75}& \textbf{35.12}& \textbf{82.43}& \textbf{68.72}& \textbf{40.25}& \textbf{32.42}& \textbf{53.40}& \textbf{52.74}& \textbf{36.81}& \textbf{17.00}& \textbf{78.59}& \textbf{45.27}& \textbf{17.53}& \textbf{13.24}& \textbf{46.82}& \textbf{22.87}\\\hline
 \end{tabular}
 % }}
  % \vspace{-3mm}
       \caption{VSG performance comparison, where ``FS'' is ``fully-supervised'' and ``WS'' is ``weakly-supervised''.}
       % \vspace{-5mm}
    \label{tab:vsg_act_cha}
\end{table*}

% \noindent \textbf{Choosing the most discriminative word.} 
\noindent \textbf{Self-weighted supervision.}
Since the visual features have higher computational complexity, we generate the positive and negative texts only by the original text (\textit{i.e.}, anchor text) without considering the video input. Since different words contribute variously to sentence understanding,
% When analyzing the sentence text, we  
we target to find the most discriminative word for better text understanding by the following loss: 
% Therefore, we define $\mathcal{L}_i$  in  \eqref{eqn:cons_act} a follows:
\small
\begin{equation}\label{component_l}
    \mathcal{L}_{CL}^i = \max ( \mathcal{L}_{cl}^{i, 1}, \mathcal{L}_{cl}^{i, 2}, \dots, \mathcal{L}_{cl}^{i, C} ). 
\end{equation}\normalsize
For $C$  losses ($\mathcal{L}_{cl}^{i,1}, \dots, \mathcal{L}_{cl}^{i,C}$), each   $\mathcal{L}_{cl}^i$ corresponds to a specific negative text, where the corresponding sentence component is changed. In  Eq. \eqref{component_l}, the maximum of these decomposed losses corresponds to the sentence component that is most clearly identified.
{Considering the significance score, we can obtain the finally video-guided self-weighted cross-modal bridging loss:}
\small
\begin{equation}\label{final}
   \mathcal{L}_{weighted} \!=\!\sum\nolimits_{i,j,m,c}\! S_1(q_i, Q_m, c)\cdot S_2(Q_m^i,Q_m^j, c)\cdot\mathcal{L}_{CL}^i. 
\end{equation}\normalsize
% \noindent \textbf{Cross-modal fusion.}
% To effectively fuse video features and text features, we introduce a cross-modal late fusion strategy for cross-modal fusion. Based on the text feature $Q$, we extract the frame-wise video features $\{v_i\}$ by affine transformation as follows:
% \small
% \begin{equation}
% \begin{aligned}
% {X}^{(l)}=\tilde{\mathbf{c}}\odot\mathrm{MLP}(\mathrm{LN}(\tilde{\mathbf{Z}}^{(l)}))+\tilde{\mathbf{Z}}^{(l)}, v_i&={W}^{(l)}\odot v_i+{B}_i,
%     ({W}_i,{B}_i)=\text{MCA}(\text{LN}({v}_i),\text{LN}(Q)),
% %     \\
% % v_i&={W}^{(l)}\odot v_i+{B}_i,
% %     \\
% %     {X}^{(l)}&=\tilde{\mathbf{c}}\odot\mathrm{MLP}(\mathrm{LN}(\tilde{\mathbf{Z}}^{(l)}))+\tilde{\mathbf{Z}}^{(l)},
% \end{aligned}
% \end{equation}\normalsize
% where $\mathrm{LN}(\cdot)$ denotes the layer normalization operation, $\mathrm{MLP}(\cdot)$ denotes the multi-layer perceptron network, and $\mathrm{MCA}(\cdot)$ denotes the multi-head cross-attention operation~\citep{vaswani2017attention}. In the output, we can obtain the affine weight $\mathbf{W}^{(l)}$ and the corresponding bias $\mathbf{B}^{(l)}$. Also, we denote the learnable per-channel scales as $\tilde{\mathbf{c}}$ and the modulated video feature as $\mathbf{X}^{(l)}$.
Since our method is plug-and-play, we borrow the cross-modal fusion module from an open-source works into our framework, which is the base version of our method. 

\section{Experiments}
% \subsection{Cross-modal Fusion for Grounding}
% \label{subsec:va_attn}
\noindent \textbf{Datasets.} For a fair comparison, we utilize the following datasets:
1) For  VSG, we utilize three datasets: ActivityNet Captions \citep{caba2015activitynet}, and Charades-STA \citep{sigurdsson2016hollywood} and {TACoS} \citep{regneri2013grounding}. 2) For  VTR, we adopt two datasets: {MSRVTT}~\citep{xu2016msr} and {LSMDC}. 3) For  VideoQA, we use two datasets: {NExT-QA} \citep{xiao2021next} and {STAR} \citep{wu2021star}. 
% \textbf{Full details are placed in  supplementary material.}

\noindent \textbf{Compared methods} For better reproducibility, we compare some open-source state-of-the-art  works from three multi-modal tasks. 1)  VTR (text-to-video retrieval and video-to-text retrieval. 2) VideoQA. 3)  VSG.
% 1)  VTR (text-to-video retrieval and video-to-text retrieval):  X-Pool~\citep{gorti2022x}, CLIP-ViP~\citep{xue2022clip}, UATVR \citep{fang2023uatvr}, T-MASS \citep{wang2024text}. 2) VideoQA: All-in-One \citep{wang2023all}, Just Ask \citep{yang2021just}, MIST \citep{gao2023mist}, HiTeA \citep{ye2022hitea}, InternVideo \citep{wang2022internvideo}, BLIP-2~\citep{li2023blip},  VideoQA-TA \cite{wu2025videoqa}. 3) VSG: 2D-TAN \cite{zhang2019learning}, MMN \cite{wang2022negative}, G2L \cite{li2023g2l}, VCA \cite{wang2021visual}, BM-DETR \cite{jung2025background}, WSTAN \cite{wang2022weakly}, CNM \cite{zhengming2022weakly}, MMDist \cite{bao2024local}, MCMT \cite{zhang2025multi}.
% SeViLA \citep{yu2023self},

\noindent \textbf{Evaluation metrics.}
For  VTR, we utilize Recall at rank $\{1,5,10\}$ (R@1, R@5, and R@10), Median Rank (MdR), and Mean Rank (MnR) for evaluating the retrieval performance. 
For  VSG, we evaluate the grounding performance by ``R@n, IoU=m'', which means the percentage of queries having at least one result whose Intersection over Union (IoU) with ground truth is larger than m. 
% In our experiments, we use $n\in\{1,5\}$ for all datasets, $m \in \{0.5,0.7\}$ for ActivityNet Captions and Charades-STA, $m \in \{0.3,0.5\}$ for TACoS.
For  VideoQA, we introduce seven metrics: temporal (Tem), causal (Cau), description (Des), interaction (Int), sequence (Seq), prediction (Pre) and feasibility (Fea). 
% In these  metrics, lower MdR and MnR denotes better performance. For the other metrics, higher value means better performance.  
Bold denotes the best performance.

\subsection{Performance Comparison}
% \vspace{-2pt}
% We conduct performance comparisons on all three datasets under both closed-set and  settings. To evaluate efficiency, we only choose the open-source compared methods  that are grouped into two categories: (i) \textbf{Fully-supervised (FS)} setting  \citep{gao2017tall,li2023g2l,liu2018attentive,li2023momentdiff,yuan2019semantic,zhang2019cross,zhang2019learning,zeng2020dense,gao2021fast,zhang2021multi,gao2021relation,wang2022negative};
% (ii) \textbf{Weakly-supervised (WS)} setting 
% \citep{chen2022explore,yang2021local,zhang2020counterfactual,wang2021visual,wang2021weakly,zheng2022weakly}.
% %, WSSTG \citep{chen2019weakly}
% For convenience, we denote our proposed ``\textbf{robust setting}'' as ``\textbf{RS}''.
Following previous open-source  methods, we directly cite the corresponding results from compared methods.  In this paper, we treat our as the plug-and-play module for state-of-the-art VLM models to improve their performance. 

\noindent \textbf{Performance comparison on the VTR task.} VTR is a challenging multi-modal task, which requires the designed model can effectively bridge the gap between videos and texts. In this paper, we consider two subtask: text-to-video  retrieval and video-to-text retrieval. Table \ref{tab:videoqa} illustrates the effectiveness of our model as the plug-and-play module for previous VTR methods. We can find that when using augmented text, all the compared methods suffer performance degradation. The core reason is that previous VTR methods pay less attention to the language input, and ignore much language information in the sentence query. By using our model as the plug-and-play module, previous method can obtain significant performance improvement.
% since our proposed model can fully mine latent language semantics.

\noindent \textbf{Performance comparison on the VideoQA task.} Similar to the VTR task, we conduct performance comparison VideoQA performance comparison. The experimental results are summarized in Tables  \ref{tab:star} and \ref{tab:nextqa}, where the performance of previous methods was unsatisfactory. 
% The key reason is that previous methods have difficulty in understanding the rewritten question. Different from them, we can explore more deep and fine-grained language information by attribute-based text reasoning.

\noindent \textbf{Performance comparison on the VSG task.} We conduct  VSG performance comparison on all three datasets  under both fully- and weakly-supervised settings. Tables \ref{tab:vsg_act_cha}  reports the quantitative comparison results. Obviously, our proposed model can help state-of-the-art VSG methods for performance improvement over all metrics on three datasets, which demonstrates the superiority of our proposed model. 

\begin{table}[t!]
% \vspace{-10pt}
% \small
% % \vspace{-10pt}
% \scalebox{0.75}{
% \setlength{\tabcolsep}{0.5mm}{
 \scriptsize
\setlength\tabcolsep{0.04cm}
\begin{tabular}{c|cccc|cccccccccccccccc}
\hline
\multirow{3}*{Model}&\multicolumn{4}{c|}{ActivityNet Captions} & \multicolumn{4}{c}{Charades-STA} \\\cline{2-9}
& R@1 & R@1 & R@5 & R@5& R@1 & R@1 & R@5 & R@5\\
 & IoU=0.3 & IoU=0.5 & IoU=0.3 & IoU=0.5& IoU=0.5 & IoU=0.7 & IoU=0.5 & IoU=0.7\\\hline
Ours(a)& 53.77& 40.28& 76.94& 72.25& 28.51& 20.34& 67.85& 38.71 \\
% w/o CEE&  \\
Ours(b) & 55.35& 42.03& 79.50& 74.91& 30.88& 23.92& 70.66& 41.58\\
Ours(c)& 57.63& 43.86& 81.34& 77.99& 32.50& 24.03& 71.76& 42.92 \\
% w/o CF & 58.42& 44.97& 80.53& 76.45& 33.28& 25.11& 72.33& 43.05\\\hline
\textbf{Ours(full)} &  \textbf{60.90}&  \textbf{46.86}& \textbf{84.39}&\textbf{ 80.62} &  \textbf{34.85}& \textbf{27.96}& \textbf{74.28}& \textbf{46.70} \\ \hline
\end{tabular}
% }}
% \vspace{-3mm}
\caption{Main ablation study for VSG  with G2L as the base model, where we remove each key individual component to show its effectiveness. 
% ``AQPL'' denotes ``Augmenting texts by Pre-trained LLMs'',  ``GPNQ'' denotes ``Generating Positive and Negative texts'', ``SE'' denotes ``Significance Estimation for Sentence Component Integration'', ``CF'' denotes ``Cross-modal Fusion'', ``Full'' denotes our full model.
}
% \vspace{-3mm}
\label{tab:main_ablation}
\end{table}

\subsection{Ablation Study and Analysis}
% we perform exhaustive ablation studies to analyze the effectiveness of each individual component of VSTMM. We run all the experiments 5 times and report the performance on average.

\noindent\textbf{Main ablation studies.} To demonstrate the effectiveness of each component in our model, we conduct ablation studies regarding the components 
% (\textit{i.e.}, Augmenting texts by Pre-trained LLMs, Generating Positive and Negative texts, Significance Estimation for Sentence Component Integration and Cross-modal Fusion) 
in Table \ref{tab:main_ablation}.  In particular, we remove each key individual module to investigate its contribution. For convenience, we design four ablation models: 1) Ours(a). We remove the  ``multi-level text augmentation'' module while keeping the other  modules. 2) Ours(b). We remove the ``attribute-based text reasoning'' module while keeping the other  modules. 3) Ours(c). We remove the ``video-guided self-weighted cross-modal bridging'' module while keeping the other  modules. 
% 4) Ours(d). We remove the ``Cross-modal Fusion'' module while keeping the other three modules.  
Besides, we use our full model as the baseline: Ours(full).
% Since we focus on the  setting, we conduct corresponding experiments in Table~\ref{tab:ActivityNet}, \ref{tab:Charades} and \ref{tab:TACoS}. 
% Based on these tables, we can observe that 
All four modules contribute a lot to the final performances on all three datasets, demonstrating their effectiveness in the  VSG task. 
% As the key module to generate texts, the ``Augmenting texts by Pre-trained LLMs'' module brings the largest improvement, showing that it can conduct effective text rewriting for cross-modal reasoning. Besides,  by ``Generating Positive and Negative texts'', we can sufficiently understand the given text with the help of positive and negative texts. Also, ``Significance Estimation for Sentence Component Integration'' improves the grounding performance in terms of all the metrics, which illustrates the effectiveness of our significance estimation strategy. For ``Cross-modal Fusion'', it also helps our model for further performance improvement.

% the uncertainty-aware OOD boundary reasoning module achieves significant performance improvement, which illustrates the effectiveness of our uncertainty score and OOD boundary reasoning. Besides, the ID-OOD boundary refinement module improves the performance over all the metrics.

% \begin{wrapfigure}{r}{77mm}
% \begin{center}
% \vspace{-7mm}
%  \includegraphics[width=0.98\linewidth]{qualitative_01.jpg}
% \end{center}
% \vspace{-5mm}
% \caption{
% \small Visualization results.
% % Jailbreak performance of different categories of malicious questions.  
% }
% \label{fig:vis}
% \vspace{-8mm}
% \end{wrapfigure}

\begin{figure}[t!]
% \vspace{-15pt}
\centering
% \subfigbottomskip=-3.6pt
% \subfigcapskip=-5.5pt
%\vspace{0.35cm}
\includegraphics[width=0.235\textwidth]{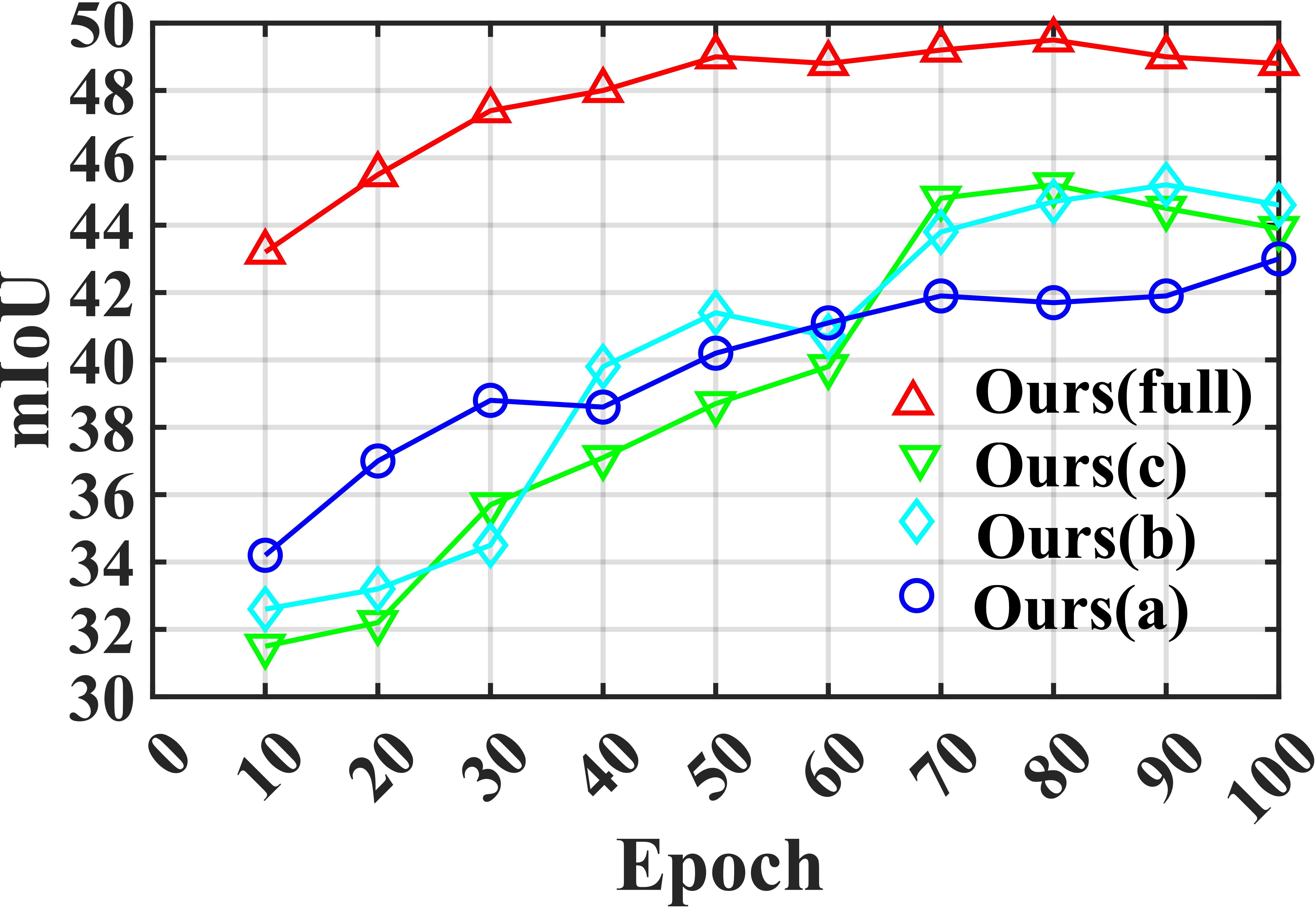} 
\hspace{-0.05in}
\includegraphics[width=0.235\textwidth]{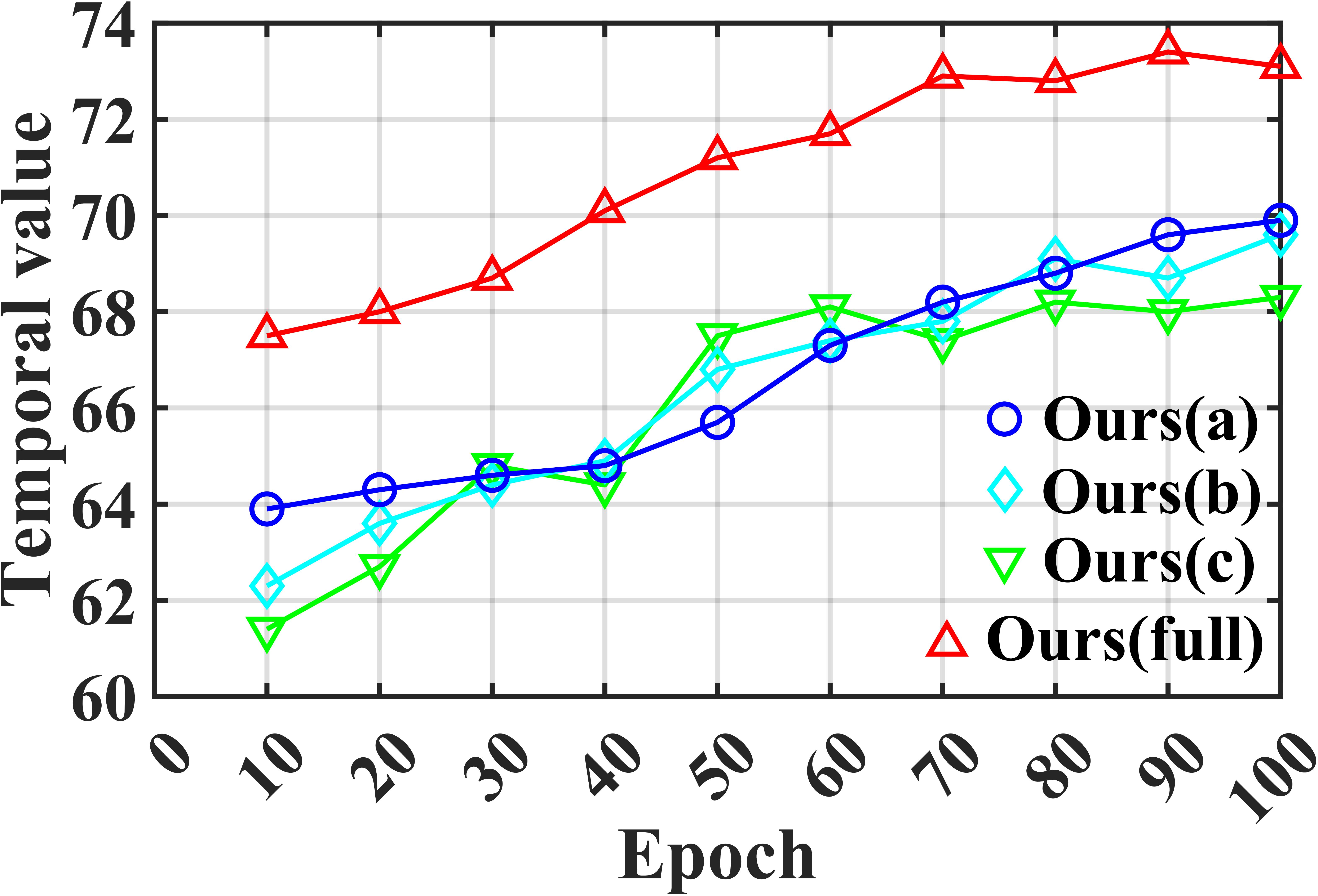} 
% \hspace{-0.1in}
% \includegraphics[width=0.33\textwidth]{msr_vtt_xunliantu_iclr_01.jpg} 
% \vspace{-0.3cm}
\caption{ Training performance of each ablation module with text augmentation on the ActivityNet Captions dataset (left, VSG), the NexT-QA dataset (right, VideoQA). }
 % \vspace{-0.8cm}
 % \vspace{-0.5cm}
\label{fig:xunliantu}
\end{figure}

% \begin{figure*}[t!]
% % \vspace{-15pt}
% \centering
% % \subfigbottomskip=-3.6pt
% % \subfigcapskip=-5.5pt
% %\vspace{0.35cm}
% \includegraphics[width=0.33\textwidth]{act_xunliantu_iclr_01.jpg} 
% \hspace{-0.1in}
% \includegraphics[width=0.33\textwidth]{next-qa_xunliantu_temporal_01.jpg} 
% \hspace{-0.1in}
% \includegraphics[width=0.33\textwidth]{msr_vtt_xunliantu_iclr_01.jpg} 
% \vspace{-0.3cm}
% \caption{\small Training performance of each ablation module with text augmentation on the ActivityNet Captions dataset (left, VSG), the NexT-QA dataset (middle, VideoQA) and the MSR-VTT dataset (right, VTR). }
%  % \vspace{-0.8cm}
%  \vspace{-0.7cm}
% \label{fig:xunliantu}
% \end{figure*}

\noindent \textbf{Training process of different ablation models.} Following \citep{lin2020weakly}, we analyze the training process and retrieval performance of different ablation models in Figure \ref{fig:xunliantu}. 
We can obtain the following representative observations:
(i) During training, Our(full) outperforms other ablation models, which further demonstrates the effectiveness of each module. 
% For example, compared to the second-best model HLGT(v), HLGT(full) improves the performance by 3.27\%.
(ii) Our(full) converges faster than ablation models,  showing that our full model is more efficient. For instance, Our(full) converges within 70 epochs, while Our(c) converges after 80 epochs. 
Thus, our full model can process these challenging datasets more efficiently.

\noindent\textbf{Effect of different augmentation methods.} In Table \ref{tab:ablation_aug_type}, we further compare our augmentation method in Section \ref{subsec:positive} with other methods (Back-translation, Paraphrasing and LLaMA-7B). Obviously, our method achieves the best performance. It is because LLaMA-7B generates some hallucination information during augmentation. Back-translation and paraphrasing only treat each word with the same significance, which means that some unimportant words have a large weight. We can evaluate the significance score of each word to better understand the text. Also, we generate attributes to reason the fine-grained semantics in the given sentence and utilize attribute sampling to purify these semantics-rich attributes for better text understanding. 
% Besides, we conduct multi-level text augmentation to fully understand the given text from word and structure levels .

% \noindent \textbf{Effect of different levels of text augmentation.} To analyze the contribution of text augmentation from different levels, we conduct the ablation study as shown in Table \ref{tab:ablation_text_aug_level}. 
% Both word-level augmentation and structure-level augmentation can significantly improve the grounding performance, showing the effectiveness of word understanding and structure reasoning for text understanding.

\begin{table}[t!]
% \vspace{-4mm}
 % \small
\centering
 % \vspace{-4mm}
 % \vspace{-2mm}
% \resizebox{0.5\textwidth}{!}{
% \scalebox{0.8}{
% \setlength{\tabcolsep}{0.8mm}{
 \scriptsize
\setlength\tabcolsep{0.06cm}
\begin{tabular}{c|ccc||c|cccccccccccc}
\hline
% \multirow{2}*{Method(\# Frames)} & \multicolumn{3}{|c|}{Without text augmentation}& \multicolumn{3}{c}{With text augmentation} \\\cline{2-7}
\multicolumn{4}{c||}{NExT-QA}&\multicolumn{5}{c}{STAR}\\
\hline 
Method  & Tem\small{$\uparrow$} & Cau\small{$\uparrow$} & Des\small{$\uparrow$} &Method& Int\small{$\uparrow$} & Seq\small{$\uparrow$} & Pre\small{$\uparrow$} & Fea\small{$\uparrow$}\\
\hline 
% All-in-One(32)     & 48.6 & 48.0 & 63.2 & 40.2& 37.9& 53.8  \\
% \textbf{+Ours(32)} & \textbf{50.1}& \textbf{51.9}& \textbf{64.7} & \textbf{48.6}& \textbf{50.2}& \textbf{61.3}  \\\hline 
% Just Ask(20)   & 51.4 & 49.6 & 63.1 & 42.7& 40.1& 54.0  \\
% \textbf{+Ours(20)} & \textbf{54.3}& \textbf{52.9}& \textbf{67.8} & \textbf{50.9}& \textbf{49.3}& \textbf{62.7 }\\\hline 
% MIST(32)     &  56.6 & 54.6 &  66.9 & 51.9& 48.2& 55.3 \\
% \textbf{+Ours(32)} & \textbf{60.3}& \textbf{56.9}&\textbf{ 69.8}& \textbf{57.2}& \textbf{55.4}& \textbf{67.9} \\\hline
% HiTeA(16)    & 58.3  & 62.4 &  75.6 & 52.2& 57.6& 59.3 \\
% \textbf{+Ours(16)}  & \textbf{62.8}& \textbf{65.7}& \textbf{77.3} & \textbf{60.4}& \textbf{63.9}& \textbf{74.9}\\\hline
% InternVideo(8)  &  58.5 & 62.5 &  75.8 & 52.9& 57.4& 70.3 \\
% \textbf{+Ours(8)}  & \textbf{62.5} & \textbf{66.3}& \textbf{76.4} & \textbf{61.8}& \textbf{59.7}& \textbf{74.5} \\\hline
% Synonym replacement&	64.4&	62.3&	71.0\\
Back-translation&	62.8&	62.7&	68.9&Contrastive loss& 61.4& 62.7& 54.9& 52.6\\
Paraphrasing&	65.3&	63.4&	72.5& Ours(w/o $S_1$)& 64.8& 68.4& 56.6& 55.0\\
 LLaMA-7B(*)  & 66.8& 65.4& 74.8 & Ours(w/o $S_2$)& 65.1& 67.2& 57.9& 54.6\\
% BLIP-2(4)  & 67.2  & 70.3 &  79.8 & 64.0& 61.9& 72.3 \\
\textbf{Ours}  &\textbf{69.2}& \textbf{70.1}& \textbf{78.4} &\textbf{Ours(full)} & \textbf{66.2}& \textbf{71.6}& \textbf{58.7}& \textbf{55.3}\\\hline
% SeViLA(4)   &  {{67.7}}  & {{72.1}} & {{82.2}} & 64.0& 66.8& 76.9 \\
% \textbf{Ours}  & \textbf{69.2}& \textbf{74.9}& \textbf{85.3} & \textbf{70.5}& \textbf{72.7}& \textbf{83.9} \\\hline
% SeViLA~\citep{yu2023self}    & 8 &  {{67.0}}  & {{73.8}} &  {{81.8}} \\
% \textbf{+Ours} & \textbf{8} & \textbf{72.8}& \textbf{76.9}& \textbf{87.2} \\\hline
% \midrule
%ViLA (4) (Ours)  &   69.1 & 73.6 & 81.3  &  \textbf{74.2} \\
% \textbf{ViLA (Ours)}  & 4 (8 to 4) &   {\textbf{71.0}} & 72.9 & {\textbf{82.7}}  \\
% \textbf{ViLA (Ours)}  & 4 (32 to 4) &   70.1 & {\textbf{73.8}} & 82.1  \\
%ViLA (Ours)  & 16 &  69.5 & 74.0 &  81.7 &  \textbf{75.0} & 188M \\
% \midrule
% %ViLA (Ours)  & 32 &  \textbf{72.3}  & \textbf{74.9} &  82.1 &  \textbf{75.6} & 188M & -\\
% \textbf{ViLA+LoRA (Ours)}  & \textbf{4} &  71.4  & 74.5 &  80.3 \\
\end{tabular}
% }}
% \vspace{-3mm}
\caption{Ablation study on different augmentation methods on NExT-QA and $\mathcal{L}_{weighted}$ in Eq. \eqref{final} on STAR for  VideoQA, where BLIP2 is the base model with \# Frames=4.  ``LLaMA-7B(*)'' means that we directly use LLaMA-7B for text augmentation. 
% VideoQA performance comparison on NExT-QA dataset, where the value means the accuracy of providing the right answer. 
% Here we measure the accuracy of choosing the right answer. 
%Our ViLA outperforms the SOTA method by {\bf 0.9-1.0\%} at 4 frames with {\bf1.45-3.04$\times$} speed up and  by {\bf1.0\%} at 8 frames with {\bf1.35$\times$} speed up.
%and push the accuracy to reach 75.5\% on this dataset at 32 frames setting. 
%And our proposed ViLA with 4 frames even achieved better performance than SeViLA with 8 frames. 
% Especially on Temporal and Causal type of questions, our ViLA (using only 4 frames) improves {\bf 3.3\%} and {\bf 1.7\%} respectively, compared with SeViLA. 
% We use \textbf{bold-face} font to indicate the best results and \underline{underline} on the second best using the same number of frames.
% ViLA using 2-frames only out-performs BLIP-2 using 4-frames by \textbf{1.3\%}.  ViLA also achieves upto {\bf 3.04$\times$} speedup. It needs to be noted that our ViLA achieves 75.1\% average accuracy with only 4 frames when we finetune LLM with LoRA~\citep{hu2021lora}.
%The comparison with SeViLA shows the effectiveness of our Frame-Prompter and QFormer-Distillation.
%\ljb{can you have the brown box for the text 'brown box' and the same for blue box as well?}
}
\label{tab:ablation_aug_type}
% \vspace{-3mm}
\end{table}

\noindent\textbf{Influence of $\mathcal{L}_{weighted}$ in Eq. \eqref{final}.} To evaluate the effectiveness of our video-guided self-weighted cross-modal bridging loss, we conduct ablation study in Table \ref{tab:ablation_aug_type}.  Obviously, directly using the contrastive loss will lead to unsatisfactory since it only construct positives and negatives from a level, we construct positives and negatives from different levels (word-level and structure-level). Comparing Ours(full) with Ours(w/o $S_1$) and Ours(w/o $S_2$), we can achieve significant performance improvement since our model can fully understand the text inputs based on $S_1$ and  $S_2$.

\section{Conclusion}
In this paper, we rethink the LLM task from the user-friendly language input perspective. 
{ We observe that many VLMs cannot fully understand the language texts. Given some texts with similar semantics and a  video, these VLMs output various  results. }
% Based on the observation that texts with similar semantics but different templates lead to various  results for the same video, we find that many VLMs cannot fully understand the language texts. 
 To this end,  we design a novel plug-and-play framework to improve the generation ability of previous methods on various text templates. 
 % We first generate diverse positive and negative texts by augment the original text from  word  and structure levels.
% To obtain diverse positive and negative texts, we augment the text by both word-level and structure-level for rewriting texts.
% To selectively integrate these generated texts, we  understand the text input from different granularity.  Moreover, a video-guided self-weighted cross-modal bridging loss is proposed to integrate these sentence components by assigning adaptive weights to these components.
Extensive experiments  show that our framework can serve as the plug-and-play module for state-of-the-art VLM works to improve their performance on various video-language tasks. 
% % In our future work, we will extend our model into more multi-modal tasks.
In our future work, we will extend our model into image-language model or video-audio model to achieve broader applicability.

% leverage some large-scale multi-modal pre-training models to rebuild our framework to achieve broader applicability and robustness.

% \section{Impact Statement}
% This paper presents work whose goal is to advance the field of Machine Learning. There are many potential societal consequences of our work, none which we feel must be specifically highlighted here.

% We present a novel framework for weakly supervised adaptive contrastive learning for multi-modal video-language tasks. Specifically, we mitigate the issue of models' deficiency in the perception of different sentence components by utilizing LLM-generated component-targeted negative samples. We additionally integrate our proposed adaptive importance estimation module to accommodate sample-wise variations in the significance of different types of sentence components. We evaluate our method across three different baselines and two different video-language joint learning tasks. In each task, our method outperforms the baseline considerably, validating the effectiveness of our proposed method.

% Bibliography is inlined for arXiv to avoid BibTeX reruns on the source .bib files.

\end{document}